\documentclass[preprint,12pt]{elsarticle}

\usepackage{amsmath, amssymb, amsthm}
\usepackage{bm}          % bold math symbols
\usepackage{booktabs}    % professional tables
\usepackage{array}       % extended column specs
\usepackage{hyperref}    % clickable refs (optional, safe to 
\usepackage{booktabs}
\usepackage{array}
\usepackage{multirow}
\usepackage{xcolor}
\usepackage{siunitx} 
\usepackage{caption}
\usepackage{subcaption}
\usepackage{amssymb}
\usepackage{amsmath}
\newtheorem*{remark}{Remark}
\usepackage[inkscapelatex=false]{svg}

\begin{document}

\begin{frontmatter}



\title{Learning Traffic Flow Dynamics with Stochastic Physics-Informed Neural Cellular Automata}

%%author[label1,label2]{}

\author[label1]{Federica Bragone} %% Author name

\affiliation[label1]{organization={Department of Decision and Control Systems, KTH Royal Institute of Technology},%Department and Organization
            city={Stockholm},
            country={Sweden}}

\author[label1]{Matthieu Barreau}

%% Abstract
\begin{abstract}
Traffic flow modeling is essential for understanding and predicting the collective dynamics of vehicles on road networks. Cellular automata provide a simple, interpretable yet powerful framework for representing these dynamics via local interaction rules, while retaining the ability to reproduce complex macroscopic traffic phenomena. However, learning local transition rules from data while preserving physically meaningful constraints remains challenging, particularly for stochastic models. In this work, we propose a physics-informed neural cellular automaton (PI-NCA) for data-driven traffic flow modeling. Building on the standard neural cellular automaton (NCA), we design a neural architecture that is physically consistent with the road topology and guarantees conservation of the total number of vehicles, thereby constraining the learned transition rules to physically admissible dynamics. We further extend this framework to stochastic dynamics by parameterizing probabilistic transition rules while preserving the same physics-informed constraints. We evaluate the proposed models on multiple traffic scenarios generated by the well-established Nagel–Schreckenberg and Kerner–Klenov–Wolf cellular automata. The results demonstrate that the PI-NCA successfully learns the dynamics of both traffic models and consistently outperforms a standard NCA, while the stochastic extension captures probabilistic transition rules without compromising the imposed physical constraints.

%As transportation demand increases, we need novel, more efficient models for simulating traffic flow. Traffic-based Cellular Automata models have been widely used to reproduce and analyze vehicles' behavior on the roads, such as the Nagel-Schreckenberg (NaSch) model, which simulates freeway traffic, and the KKW model, which simulates more complex traffic dynamics based on three-phase traffic theory. Since their first appearance, CA models have evolved and improved, and with growing interest in Machine Learning, researchers have combined them with data-driven models. A Neural Cellular Automaton (NCA) illustrates how existing models can be combined to create a powerful predictor of the future states of the systems under consideration. In this work, we propose a classical NCA model to predict vehicle states on a one-lane circular road, and extend it to a Physics-Informed Neural Cellular Automata (PI-NCA) to predict vehicle velocities and explicitly update their positions. We test the model on both deterministic and stochastic datasets generated using the NaSch and the KKW-1 models, demonstrating the proposed PI-NCA's ability to learn causal and random rules.
\end{abstract}


% \begin{highlights}
% \end{highlights}

%% Keywords

\end{frontmatter}

\section{Introduction}
	\label{sec:introduction}

Accurate traffic models are fundamental to simulating, forecasting, assessing, and controlling transportation systems. However, obtaining a high-fidelity simulator is challenging, as traffic flow is a dynamical system arising from interactions among individual vehicles and the surrounding road traffic conditions. Classic approaches consider various scales from microscopic modeling to macroscopic, with multiple parameters representing physical quantities \cite{ferrara2018freeway}. However, determining these parameters is cumbersome because they are not readily measurable. In recent years, a data-driven paradigm has emerged to free from these parameter dependencies \cite{xing2023overview,rowan2025systematic,borkar2025comparative}. For instance, Machine Learning (ML) methods have gained increasing recognition for modeling traffic dynamics, including recurrent architectures like Long Short-Term Memory (LSTM) \cite{wang2019foreseeing, culita2020hybrid}, Convolutional Neural Networks (CNNs) \cite{de2020spatio, di2019traffic}, and Graph Neural Networks \cite{agafonov2020traffic}. Gomes et al. \cite{gomes2023survey} provide a broader overview of ML approaches applied to traffic flow prediction.

Beyond classical data-driven approaches, Cellular Automata (CA) provide a relatively simple and computationally efficient way to model microscopic traffic. Traffic CA discretize space and time, where vehicle motion is governed by local physics-inspired transition rules applied simultaneously across the road. Even with simple rules, these models can produce non-trivial behaviors, including free flow, congestion, and wave dynamics. A key achievement in this field was the development of the Nagel-Schreckenberg (NaSch) model \cite{nagel1992cellular}, which simulates freeway traffic using rules for acceleration, deceleration, random deceleration, and movement. This CA was the first to exhibit the empirically validated meta-stability phenomenon, in which the same road configuration could lead to either free flow or congestion \cite{appert2001introduction}. Another well-known example is the Kerner-Klenov-Wolf Traffic Cellular Automaton (KKW-TCA) model \cite{kerner2002cellular}, which represents the three phases of traffic flow, as described by Kerner in \cite{kerner1999physics, kerner2004three, kerner2009introduction} with a limited number of simple rules. More recent traffic CA have extended these approaches to include more heterogeneous conditions. For example, Vranken et al. \cite{vranken2021adapting} developed a traffic CA to model traffic consisting of human-driven, automated, and communicating automated vehicles. Jing et al. \cite{jiang2021cellular} modeled mixed traffic flow by analyzing the driving behavior of connected automated vehicle platoons. 

The main issue with CA is the complexity of the tuning tasks required to faithfully reproduce observations. Therefore, some works included data-driven models; for example, Shang et al. \cite{shang2022data} used LSTM and SVM models to learn car following and lane changing behavior, respectively, from observed vehicle data and incorporated these learned behavioral components into a traffic CA. Interestingly, CA and CNNs share a local, spatially invariant structure across the spatial domain \cite{lecun2002gradient, lecun2015deep, krizhevsky2012imagenet}, suggesting that CA rules could be learned. Indeed, Gilpin \cite{gilpin2019cellular} showed that any CA rule set can be represented and learned by a CNN architecture, highlighting that networks trained on simpler CAs develop hierarchical internal structures, whereas more complex CAs yield shallower networks. In \cite{silverman2019convolutional}, CA trajectories are treated as images, and CNNs are used to categorize their emergent pattern behavior into Wolfram classes \cite{wolfram2002nks} automatically. In \cite{aach2021generalization}, the authors show how CNNs can learn unseen rule sets for Conway's Game of Life. In \cite{rollier2025convolutional}, the authors use CNNs to identify behavioral classes in CA. Related research on Neural Cellular Automata (NCA) has further demonstrated how learnable local rules can generate and reproduce complex spatiotemporal patterns \cite{mordvintsev2021texture, mordvintsev2022growing, randazzo2023growing}. 
A combination of CA and neural networks has also been applied to traffic modeling. Chai and Wong \cite{chai2014neural} introduced a traffic model that combines a rule-based traffic CA with a pretrained artificial neural network to support vehicle lane-choice decisions. More recently, Yang and Jerath \cite{yang2024deep} used a CA traffic model to generate training data for a CNN--LSTM architecture for traffic flow prediction. In \cite{ziabari2026labeled}, the authors combined a labeled traffic CA based on three-phase traffic theory with Graph Neural Networks to classify traffic phases for traffic control applications. 

These developments show promising results in how neural networks can complement CA to alleviate the heavy parameter-tuning task. However, in traffic, CA have mainly been used to generate data for training prediction or classification models, or used neural networks to learn only one specific part of the CA model. In contrast, less attention has been given to learning the local traffic transition rules directly from state trajectories and using the learned rules to simulate traffic dynamics. The ability of learned local rules to reproduce realistic traffic dynamics, such as preserving the number of vehicles, generalizing to unseen traffic conditions, and capturing both deterministic and stochastic behavior, has not been explored in depth. Building on these observations, we investigate whether NCA can learn microscopic traffic transition rules directly from simulated trajectories and reproduce vehicle dynamics and corresponding macroscopic traffic behavior. 

To achieve this goal, we propose two models. The first is directly inspired by NCA with a specific neural architecture to represent realistic local behavior, in which a one-dimensional CNN learns the local transition rule and directly predicts the next traffic state. The second one, defined as Physics-Informed Neural Cellular Automata (PI-NCA), instead predicts the vehicles' velocities and explicitly updates their positions following the conservation law. We then extend the PI-NCA formulation to stochastic traffic dynamics by predicting categorical distributions over possible next vehicle velocities. In this way, the model learns and represents the probabilistic nature of the underlying transition rule.
We evaluate the learned models via autoregressive rollout at both the microscopic and macroscopic levels, considering predictive accuracy, vehicle conservation, generalization to unseen traffic configurations, and reproduction of characteristic traffic flow behavior as metrics. We assess the framework on data generated using NaSch dynamics and, furthermore, on more complex interactions using the KKW-1 model \cite{kerner2002cellular}, which introduces synchronization-based interactions and therefore provides more complex reference dynamics than the NaSch model.

%%%%%%%%%%%%%%%%%%%%%%%%%%%%%%%%%%%%%%%%%%%%%%%%%%%%%%%%%%%%%%%%%%
\section{Preliminaries and Problem Statement}
\label{sec:background}

\subsection{Traffic Flow Modelling}
\label{subsec:traffic_flow_modelling}
Traffic flow models provide mathematical representations of traffic system evolution and can be formulated at different levels of aggregation. Three commonly considered scales are macroscopic, mesoscopic, and microscopic models. Macroscopic models treat traffic as a continuous flow, using variables such as density, mean speed, and traffic flow \cite[Ch.~3]{ferrara2018freeway}, whereas microscopic approaches simulate individual vehicles and their interactions \cite[Ch.~5]{ferrara2018freeway}. Mesoscopic models, instead, are an intermediate level that combine aggregate and individual characteristics of the traffic flow \cite[Ch.~5]{ferrara2018freeway}.

At the macroscopic level, traffic conditions can change across both space and time and can be described by the density $\rho(x,t)$, average velocity $v(x,t)$, and traffic flow $q(x,t)$. These quantities are related through the hydrodynamic equation $q(x,t) = \rho(x,t) v(x,t)$.
The evolution of traffic density on a single lane is governed by vehicle conservation, from which it derives the 1D-conservation equation expressed as
\begin{equation}
    \frac{\partial \rho(x,t)}{\partial t} + \frac{\partial q(x,t)}{\partial x} = 0.
    \label{eq:conservation_equation}
\end{equation}
The problem, however, is not well-posed with more unknowns than equations. A practical workaround is to consider the first-order assumption that the flow is a function of the density: $q=Q(\rho)$. This results in the first-order conservation law
\begin{equation}
    \frac{\partial \rho(x,t)}{\partial t} + Q'(\rho)\frac{\partial \rho(x,t)}{\partial x} = 0,
\end{equation}
which is the basis of first-order macroscopic traffic flow models, such as the Lighthill-Whitham-Richards (LWR) model \cite{richards1956shock,lighthill1955kinematic}.

The relationship $q=Q(\rho)$ is commonly represented through the fundamental diagram, which describes how traffic flow varies with vehicle density. At low densities, vehicles can travel at higher velocities, and flow increases. As density increases, velocities decrease, and interactions between vehicles become limiting. When the system reaches a maximum-flow region, congestion typically occurs, with lower velocities and reduced traffic flow. Classical fundamental diagrams are then either quadratic \cite{greenshields1935study}, piecewise linear \cite{DAGANZO1994269}, or exponential \cite{NEWELL1993281}. These models are good approximations at low and high densities; however, at intermediate densities, traffic flow is not determined solely by density \cite{delle2021three}. In three-phase traffic theory, these densities refer to synchronized traffic, and flow is generally represented by a two-dimensional region in the flow-density plane, allowing a range of traffic states rather than a unique flow for each density \cite{kerner2012complexity, treiber2010three}.  In particular, multiple traffic states with different average velocities and flows may coexist at the same time or at similar densities, especially during transitions from free flow to congestion. This behavior is associated with metastability and the coexistence of traffic states, and it has been observed in stochastic traffic models \cite{schadschneider2000statistical, appert2001introduction, kanai2005stochastic}.
Moreover, the closure equation $q=Q(\rho)$, even in free or congested regimes, is in general not known: it can vary depending on the road topology (curvature, number of lanes...) \cite{taillanter2024structure} and the vehicle type (car, truck...) \cite{Puppo2016Heterogeneous,marques2021vehicle}, making it very challenging to define in practice \cite{qu2017stochastic}.

In addition to local macroscopic quantities, traffic conditions can be described across the entire road with global metrics.
In the present work, the traffic system consists of a homogeneous single-lane road of length $L$ where $N$ vehicles are traveling with a global density and average velocity defined as
\begin{equation}
    \rho_{\mathrm{global}} = \frac{N}{L} = \int_0^L \frac{\rho(x,t)}{L} dx, \qquad \bar{v}(t) = \frac{1}{N}\sum^{N}_{n=1}v_n(t),
\end{equation}
where $v_n(t)$ is the velocity of vehicle $n$. Therefore, the global density describes the overall traffic concentration, whereas local variables capture spatial variations in traffic conditions.

\subsection{Traffic Flow Modeling Using Cellular Automata}
\label{subsec:CA_traffic_flow}

In this subsection, we first define a traffic cellular automaton and then describe the Nagel–Schreckenberg Model in more detail, as it will serve as the backbone of this work.

\subsubsection{Traffic Cellular Automata}
Cellular automata are discrete dynamical systems defined on a regular lattice. The system evolves in discrete time, and each cell updates its internal state following a local transition rule that depends on a finite spatial neighborhood. Classical developments in CA trace back to von Neumann \cite{v1966theory} and were later generalized and studied extensively by Wolfram \cite{wolfram1983statistical, wolfram2002nks}. A key property of CA is that simple local rules can give rise to complex behavior at larger spatial and temporal scales. 

\begin{figure}
    \centering
    \includegraphics[width=0.85\linewidth]{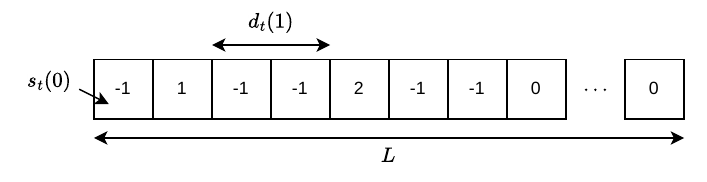}
    \caption{Traffic Cellular Automaton. Each cell contains the velocity of the car ($0$, \dots, $v_{\max}$) if there is a vehicle or $-1$ if it is empty.}
    \label{fig:TCA}
\end{figure}

In traffic applications, CA models provide a microscopic representation because vehicles move locally and their interactions can be modeled using discrete rules.
 We consider a homogeneous single-lane road represented by the one-dimensional lattice $\Omega = \{0,1,\dots, L-1\}$, where $L$ is the number of road cells, as shown in Figure~\ref{fig:TCA}. Each cell is either empty or occupied by one vehicle, and its state at time $t$ is represented by 
 \begin{equation}
 s_t(i) \in V=\{-1, 0, \dots, v_{\max}\},
 \end{equation}
 where $s_t(i)=-1$ denotes an empty cell and $s_t(i)=v_t(i)\geq 0$ represents a vehicle with velocity $v_t(i)$. The complete traffic configuration is therefore 
 \begin{equation}
     \mathbf{s}_t = \left(s_t(0),s_t(1),\dots,s_t(L-1)\right) \in V^L.
 \end{equation}
 For an occupied cell $i$ at time $t$, the distance to the next vehicle in the direction of travel is defined as
 \begin{equation}
     d_{t}(i) = \min \{d\in\{1, \dots, L\}: s_{t-1}((i+d) \bmod L) \geq 0\}.
 \end{equation}
 Consequently, $d_{t}(i)-1$ gives the number of empty cells available between the vehicle at $i$ and the vehicle immediately ahead. In this work, we consider periodic boundary conditions, meaning the road forms a closed ring and positions are evaluated modulo $L$. Therefore, the number of vehicles remains constant throughout evolution.

For a deterministic CA, the next state of cell $i$ is determined by a local function $f$ such as:
\begin{equation}
    s_t(i) = f\left(s_{t-1}(i-r),\dots, s_{t-1}(i), \dots, s_{t-1}(i+r)\right)
    \label{eq:predefined_local_function}
\end{equation}
where $r$ denotes the maximum number of cells around cell $i$ available for the local CA rule. In traffic models, the rule usually determines how vehicles adapt their velocities to nearby traffic conditions, such as their current velocity, the available distance ahead, and the velocities of neighboring vehicles. These velocities determine vehicle displacement and produce the next road configuration. By applying the same local rule throughout the road, the complete traffic state is updated from $\mathbf{s}_{t-1}$ to $\mathbf{s}_t$.
The resulting framework can reproduce macroscopic traffic patterns generated from local vehicle dynamics.
\begin{remark}
    The local, spatially invariant structure of CA directly corresponds to Convolutional Neural Networks (CNNs). A CA applies the same transition rule $f$ to local neighborhoods at every spatial position, while a convolutional layer applies a shared kernel across local regions of its inputs. This relation motivates the use of convolutional architectures to replace a predefined CA transition rule with a rule learned from data, laying the foundation for the Neural Cellular Automata defined in Section~\ref{sec:methodology}.
\end{remark}

Although traffic CA is used to study microscopic vehicle interactions, repeated application of their local rules can reproduce collective behavior such as free-flow, congestion, and stop-and-go waves \cite{maerivoet2005cellular, nagel1992cellular}. This demonstrates that complex macroscopic behavior can emerge from relatively simple local rules \cite{wolfram1983statistical, wolfram2002nks}. Moreover, microscopic traffic CA can be related to macroscopic conservation-law descriptions. In particular, previous work has shown connections between traffic CA and kinematic-wave models governed by conservation equations of the form as in \eqref{eq:conservation_equation} \cite{daganzo2006traffic}. Therefore, traffic evolution can be described in complementary terms by microscopic local vehicle updates and macroscopic density--flow dynamics.

Traffic CA may also contain stochastic transition rules to represent variability in vehicle behavior. In this case, the vehicle's updated velocity is not uniquely determined by the current traffic configuration. Indeed, for each vehicle located at cell $i$ at time $t-1$, the velocity at time $t$ can be considered as a discrete random variable taking values in $v_t(i) \in \{0, \dots, v_{\max}\}$.

The transition is therefore governed by a categorical distribution over the admissible velocities, corresponding to the multi-outcome generalization of a Bernoulli distribution. Consequently, the same local traffic conditions may produce different vehicle velocities across different realizations. After sampling the velocities, the vehicles are displaced, and the resulting road configuration is determined.

\subsubsection{Nagel--Schreckenberg Model}
\label{subsec:nasch_model}

\begin{figure}
    \centering
    \includegraphics[width=0.85\linewidth]{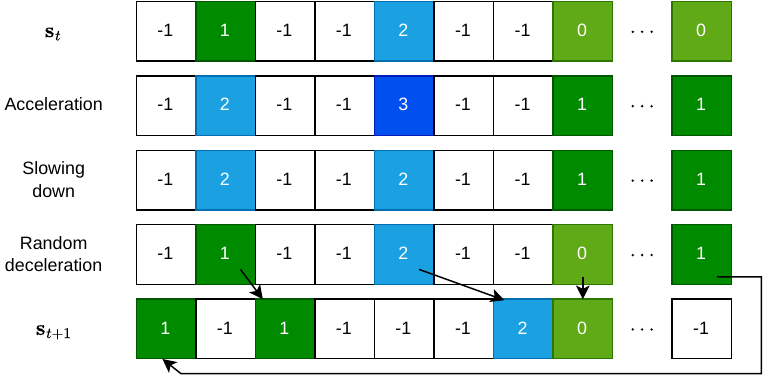}
    \caption{An example of a one-step transition for the Nagel--Schreckenberg TCA. Colors indicate the vehicle's speed in the corresponding cell. Each line corresponds to an operation: acceleration, slowing down, random deceleration, and movement to get from $\mathbf{s}_t$ to $\mathbf{s}_{t+1}$.}
    \label{fig:NaSch}
\end{figure}

The Nagel--Schreckenberg (NaSch) model is a microscopic CA model for single-lane traffic flow \cite{nagel1992cellular}. Its traffic configuration evolution, introduced in Section \ref{subsec:CA_traffic_flow}, follows four sequential update rules. For each vehicle located at cell $i$ at time $t-1$, the velocity is first updated with acceleration and collision avoidance, followed by stochastic deceleration and vehicle movement.

The update of a vehicle's velocity is carried through intermediate stages, as shown in Figure~\ref{fig:NaSch}, which we define as $v_t^{(1)}(i), v_t^{(2)}(i), v_t^{(3)}(i)$, before the vehicle moves.
\\
\textbf{Acceleration.} The vehicle's velocity increases by one unit, up to the maximum velocity,
\begin{equation}
    v_t^{(1)}(i) = \min(v_{t-1}(i) +1, v_{\max}).
    \label{eq:nasch_acceleration}
\end{equation}
\\
\textbf{Slowing down.} The velocity is restricted by the available cells in front of the vehicle,
\begin{equation}
    v_t^{(2)}(i) = \min\left(v_t^{(1)}(i), d_{t-1}(i)-1\right).
    \label{eq:nasch_slowdown}
\end{equation}
\\
\textbf{Random deceleration.} With probability $p$, the velocity is decreased by one unit,
\begin{equation}
    v_t^{(3)}(i) =
    \begin{cases}
      \max(v_t^{(2)}(i)-1, 0), & \text{with probability } p, \\  
      v_t^{(2)}(i), & \text{with probability } 1-p.
    \end{cases}
    \label{eq:nasch_random}
\end{equation}
The result of these three steps defines the velocity used for the transition to time $t$, $v_t(i):= v_t^{(3)}(i)$.
\\
\textbf{Movement.} A vehicle at cell $i$ at time $t-1$ is displaced according to its updated velocity, leading to cell $j$ at time $t$:
\begin{equation}
    j = (i + v_t(i)) \bmod L.
\end{equation}
The resulting state of each cell is given by
\begin{equation}
    \label{eq:operator_M}
    s_t(j) = 
    \begin{cases}
        v_t(i), & \text{if cell } j \text{ is occupied by the vehicle moving from cell } i, \\
        -1, & \text{otherwise}.
    \end{cases}
\end{equation}
We refer to this last stage with the operator $\mathcal{M}$ as $\mathbf{s}_t = \mathcal{M}(\mathbf{v}_t^{(3)})$.
The updated road configuration is obtained by applying these rules synchronously to all occupied cells. When $p=0$, the model is deterministic.

\begin{figure}[ht]
     \centering
     \begin{subfigure}[b]{0.49\textwidth}
         \centering
         \includegraphics[width=\textwidth]{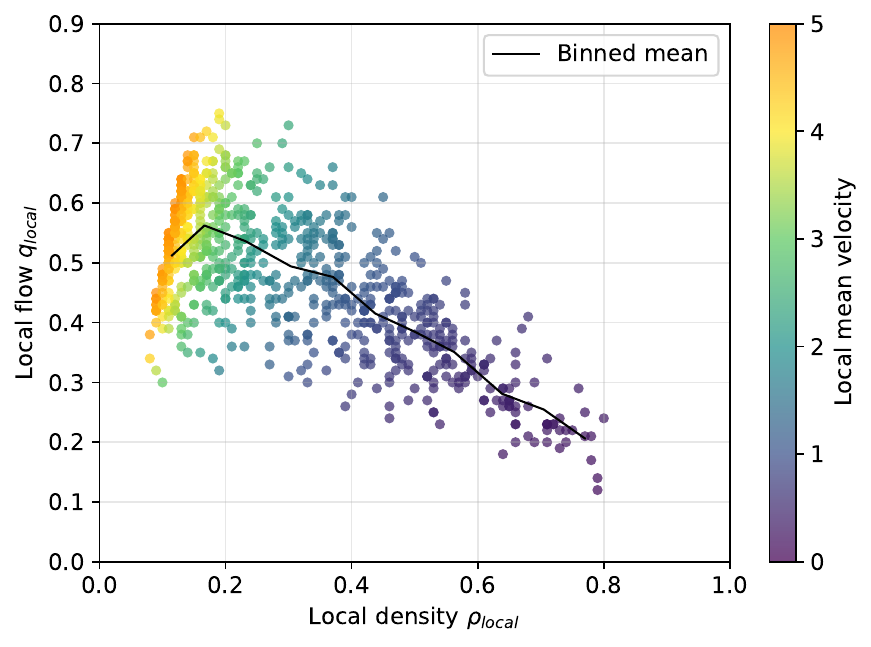}
         \caption{Local fundamental diagram}
         \label{fig:local_fd_nasch}
     \end{subfigure}
     \hfill
     \begin{subfigure}[b]{0.49\textwidth}
         \centering
         \includegraphics[width=\textwidth]{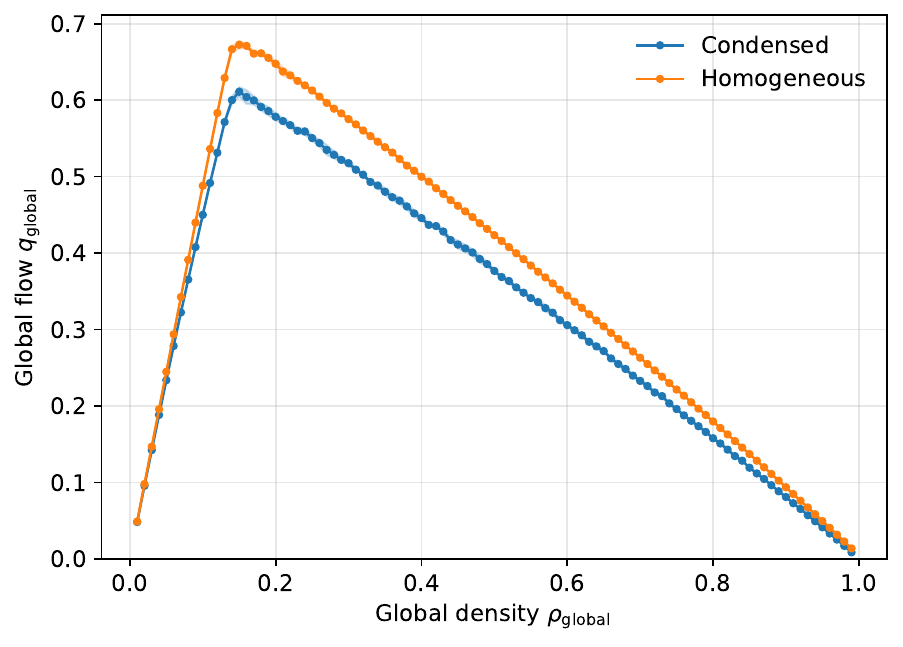}
         \caption{Global fundamental diagram}
         \label{fig:global_fd_nasch}
     \end{subfigure}
\caption{Fundamental diagrams for the NaSch model. In (a), a point represents a local pair density--flow while the color refers to the vehicle velocity. In (b), two initial conditions have been tested, producing free-flow and congested regimes.}
\label{fig:fds_nasch}
\end{figure}

The macroscopic behavior emerging from these microscopic rules can be characterized by the fundamental diagram introduced in Section~\ref{subsec:traffic_flow_modelling}.
Figure~\ref{fig:fds_nasch} shows both the local and the global density--flow relations. The local fundamental diagram, shown in Figure~\ref{fig:local_fd_nasch}, is obtained from a single simulation with $L=100$ over 5000 time steps using $v_{\max}=5$ and $p=0.2$, with local traffic quantities measured using virtual detectors over fixed spatial and temporal aggregation windows. It illustrates the range of local traffic states encountered during the simulation, with higher velocities occurring mostly at lower local densities and decreasing as traffic becomes more congested.

The global representation in Figure~\ref{fig:global_fd_nasch} summarizes traffic behavior across different vehicle densities, while the local representation captures spatial and temporal variations in traffic evolution. 
We obtain the global fundamental diagram on a road of length $L=1000$ using two initial traffic configurations: the condensed, where the vehicles are initially placed in a compact cluster, and the homogeneous, where they are distributed approximately uniformly along the road. For each density, we average the initializations over 1000 time steps and across 5 independent stochastic realizations. The simulations use a random deceleration probability of $p=0.1$. The resulting global fundamental diagram shows flow increasing at low densities until it reaches a maximum, then decreasing as density increases, consistent with the classical piecewise-linear formulation \cite{DAGANZO1994269}. The two initial configurations produce distinct flow values over a range of intermediate densities, with the homogeneous initialization reaching a higher maximum flow than the condensed initialization. At higher densities, the two flows gradually converge, respecting the multiple flow values for a given density at intermediate densities (synchronized flow).

\subsection{Problem Formulation}
\label{subsec:problem_formulation}

We consider a dataset $\mathcal{D}$ consisting of $M$ simulated traffic trajectories generated from different initial conditions,
\begin{equation}
    \mathcal{D} = \left\{X_1, \dots, X_M \right\}.
\end{equation}
Each trajectory is a sequence of road states in the form
\begin{equation}
    X = \left(\mathbf{s}_0, \mathbf{s}_1, \dots, \mathbf{s}_T\right)
\end{equation}
where $\mathbf{s}_t\in V^L$ denotes the traffic state at time $t$. We obtain each trajectory by evolving a reference traffic CA from an initial configuration. The training set consists only of simulated trajectories without the local transition rules.

Given $\mathcal{D}$, we want to address two main problems. First, we investigate how to learn deterministic local traffic rules from observed traffic trajectories while incorporating physical structure to make the learned dynamics interpretable and preserve vehicle conservation. To address this problem, we combine NCA with a physics-informed formulation that exploits the local structure of traffic CA while explicitly representing vehicle motion. 

Second, we explore whether the inherent stochasticity of the reference traffic dynamics can also be learned without losing physical interpretability of the PI-NCA formulation. 
Therefore, the final model aims to learn both the deterministic local traffic dynamics and the uncertainty arising from stochastic vehicle behavior.

%%%%%%%%%%%%%%%%%%%%%%%%%%%%%%%%%%%%%%%%%%%%%%%%%%%%%%%%%%%%%%%%%%

    \section{Methodology}
\label{sec:methodology}

\subsection{Neural Cellular Automata}
\label{subsec:cca}

The proposed transition model is formulated as an NCA and implemented using a one-dimensional CNN. The architecture is based on partial knowledge of the underlying traffic dynamics. As described in Section~\ref{subsec:CA_traffic_flow}, a traffic CA evolves by applying a local transition function $f$ to each cell's neighborhood. The specific transition function is assumed unknown during learning; however, we know that local interactions govern the traffic dynamics and that the same transition rule applies at all spatial positions. We preserve this structure and replace the predefined local rule $f$ from Equation~\eqref{eq:predefined_local_function} with a learnable convolutional mapping. As proposed in the NaSch model, a vehicle slows down (step~2) if another vehicle is within its speed range. Therefore, in the worst case, the new location of a vehicle in cell $i$ depends only on whether there is a vehicle in cells $i+1$ to $i+v_{\max}$. The local update rule of cell $i$ should then only consider cells from $i-v_{\max}$ to $i+v_{\max}$, by symmetry. 

Consider a multi-layer CNN to model the update rule. The aforementioned property suggests that the first convolution should use a kernel of size $2v_{max}+1$ (circular) with multiple output channels.
We then process the resulting local representation with a sequence of pointwise convolutional layers (kernel size 1) followed by ReLU activation. Since a $1$-convolution applies the same transformation independently at every spatial position, these layers can be interpreted as a shared multilayer perceptron that models the nonlinear function $f$ defined in \eqref{eq:predefined_local_function} and acts on the local features extracted by the first convolution. The resulting network learns a common local transition function that is applied across the entire road.

The network's forward pass is therefore given by:
\begin{align}
    & \mathbf{H}_{[1]} = \sigma\left(\mathbf{W}_{[1]} *_{\mathrm{circ}} \mathbf{s}_{t-1} + \mathbf{b}_{[1]}\right),
    \label{eq:h1}
    \\[4pt]
    & \mathbf{H}_{[\ell]} = \sigma\left(\mathbf{W}_{[\ell]} *_1 \mathbf{H}_{[\ell-1]} + \mathbf{b}_{[\ell]}\right), \qquad \ell=2 \dots H,
    \label{eq:hpw}
    \\[4pt]
    & \mathbf{Y}_t = \mathbf{W}_{\mathrm{out}} *_1 \mathbf{H}_{[H]} + \mathbf{b}_{\mathrm{out}},
    \label{eq:out}
\end{align}
where $*_{\mathrm{circ}}$ is circular convolution, $*_1$ denotes a pointwise convolution with kernel size one, and $\sigma(\cdot)$ is the ReLU activation function.
Circular padding is used to preserve the periodicity of the simulated road \cite{goodfellow2016deep}.

\begin{figure}
    \centering
    \includegraphics[width=1.0\linewidth]{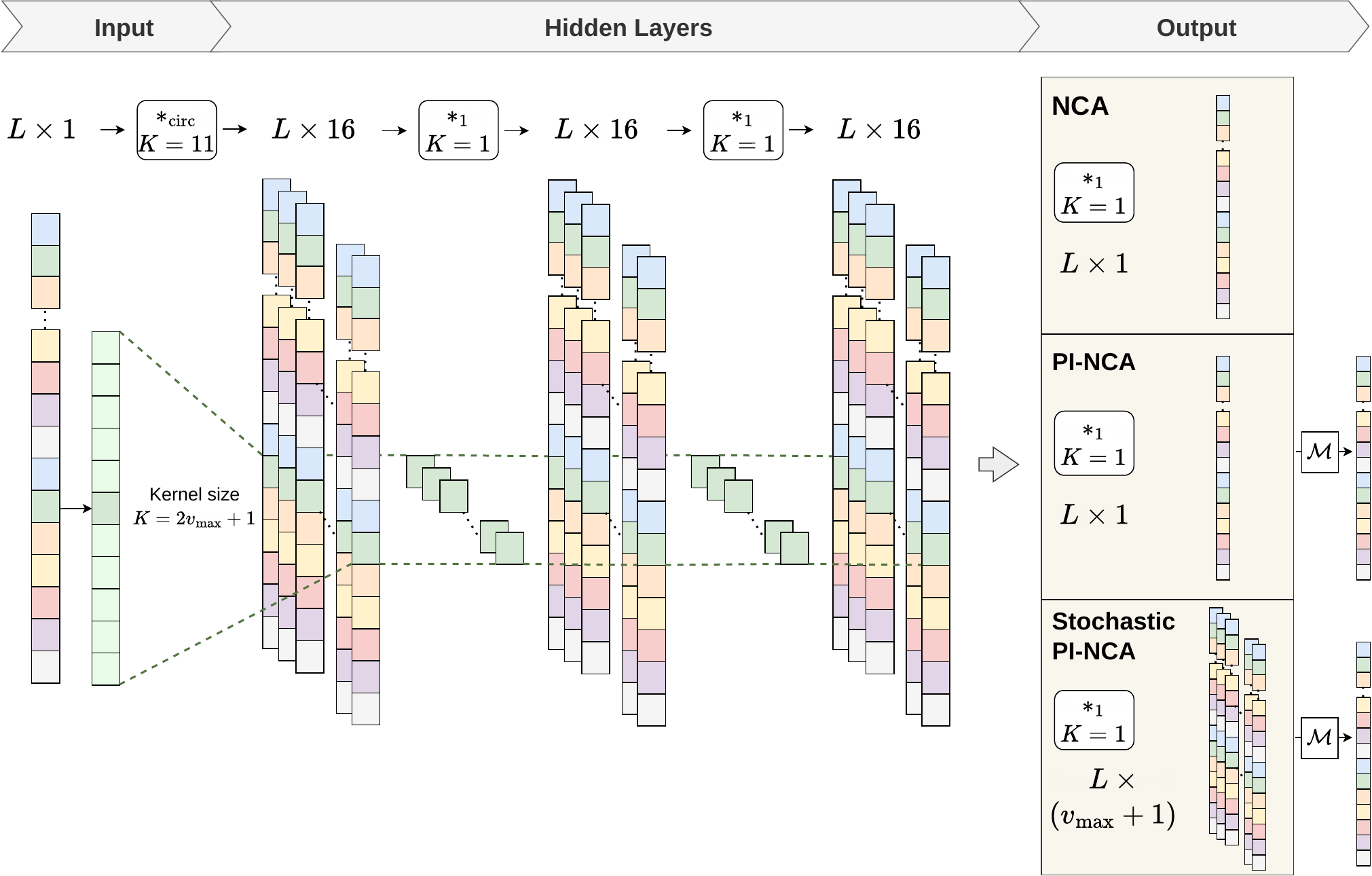}
    \caption{Convolutional Neural Network structure for NCA, PI-NCA, and Stochastic PI-NCA with 2 hidden layers, $16$ channels, and $v_{max} = 5$. The operator $\mathcal{M}$ represents the movement operation as described in Equation~\ref{eq:operator_M}.}
\label{fig:cnn_architecture_nca}
\end{figure}

Figure \ref{fig:cnn_architecture_nca} summarizes the shared convolutional architecture used by the proposed formulations. All models receive the previous road configuration $\mathbf{s}_{t-1}$ as a single-channel input and use the same local feature-extraction structure. They differ mainly in the interpretation and dimensionality of the output.

\subsection{Deterministic NCA Formulations}
\label{subsubsec:deterministic_cca}

We consider two formulations for learning deterministic traffic dynamics, as summarized in Table~\ref{tab:deterministic_models}. They share the same local convolutional structure but differ in the quantity predicted by the network and in how the subsequent traffic state is constructed.

\begin{table}[ht]
\centering
\renewcommand{\arraystretch}{1.35}
\caption{Overview of the two deterministic variants: NCA and PI-NCA.}
\label{tab:deterministic_models}
\begin{tabular}{@{} l c c c c @{}}
\toprule
\textbf{Property} & \textbf{NCA} & \textbf{PI-NCA} 
\\
\midrule
Input states & $\mathbf{s}_{t-1}$ & $\mathbf{s}_{t-1}$ 
\\
Input channels $C$ & $1$ & $1$ 
\\
Prediction target & State at $t$ & Velocity at $t$
\\
Loss & MSE & Masked MSE
\\
Loss evaluated on & All cells & Occupied cells
\\
Explicit update & No & Yes 
\\
\bottomrule
\end{tabular}
\end{table}

\paragraph{\textbf{NCA: Direct State Prediction}}

In the first formulation, denoted as NCA, the network directly approximates the local CA transition and predicts the complete traffic state,
\begin{equation}
\mathbf{Y}_t = \mathcal{F}_{\theta}^{\text{NCA}}(\mathbf{s}_{t-1}).
\end{equation}
Because the empty-cell state is represented by $-1$, the network output is transformed as
\begin{equation}
    \hat{\mathbf{s}}_t = \mathrm{ReLU}\left(\mathbf{Y}_t+\mathbf{1}\right) - \mathbf{1},
    \label{eq:state_activation}
\end{equation}
The model is trained by minimizing the Mean Squared Error (MSE) over all road cells,

\begin{equation}
    \mathcal{L}_{\text{NCA}}(\theta) = \frac{1}{L}\sum_{i=0}^{L-1}\left(\hat{s}_t(i)-s_t(i)\right)^2.
    \label{eq:state_loss}
\end{equation}
During training, the continuous output $\hat{\mathbf{s}}_t$ is retained so that the model can be optimized via gradient-based learning.
During autoregressive inference, however, the prediction must represent a valid discrete CA state before being used as input at the next time step. We therefore use
\begin{equation}
    \hat{\mathbf{s}}^{\text{disc}}_t = \mathrm{clip}\left(\mathrm{round}\left(\hat{\mathbf{s}}_t\right), -1, v_{\max}\right).
    \label{eq:state_rollout}
\end{equation}
where $\mathrm{clip}$ operation restricts every value of $\mathrm{round}\left(\hat{\mathbf{s}}_t\right)$ to lie within the interval $[-1, v_{\max}]$, and $\mathrm{round}$ converts the continuous prediction to the nearest integer.

\paragraph{\textbf{PI-NCA: Velocity Prediction with Explicit Update}}
The direct state NCA must learn both the vehicle dynamics and the resulting spatial movement from data. The second formulation, denoted as PI-NCA, includes prior knowledge that the traffic state evolves through vehicle motion. The neural network, therefore, predicts vehicle velocities, while an explicit update computes the corresponding displacement.
Let $\mathcal{C}_{t-1} = \left\{i\in\Omega: s_{t-1}(i)\geq0\right\}$ denote the set of occupied cells at time $t-1$. For each occupied cell, $i\in \mathcal{C}_{t-1}$, let $v_t(i)$ denote the reference velocity at time $t$ of the vehicle located at cell $i$ at time $t-1$, and $\hat{v}_t(i)$ be its velocity predicted by PI-NCA. The velocity prediction is
\begin{equation}
\mathbf{Y}_t=\mathcal{F}_{\theta}^{\text{PI}}(\mathbf{s}_{t-1}),
\end{equation}
with a ReLU output activation to enforce non-negative velocities
\begin{equation}
\hat{\mathbf{v}}_t = \mathrm{ReLU}\left(\mathbf{Y}_t\right).
    \label{eq:velocity_activation}
\end{equation}
Training is performed only over occupied cells, using a masked MSE,

\begin{equation}
    \mathcal{L}_{\mathrm{vel}}(\theta) = \frac{1}{|\mathcal{C}_{t-1}|}\sum_{i\in\mathcal{C}_{t-1}}\left(\hat{v}_t(i)-v_t(i)\right)^2.
    \label{eq:velocity_loss}
\end{equation}
As in the direct-state formulation, continuous predictions are used during training. During autoregressive inference, the predicted velocities are mapped to the admissible discrete velocity range,
\begin{equation}
    \hat{v}_t^{\text{disc}}(i) = \mathrm{clip}\left(\mathrm{round}\left(\hat{v}_t(i)\right), 0, v_{\max}\right), \qquad i\in\mathcal{C}_{t-1}.
    \label{eq:velocity_discretisation}
\end{equation}
Each vehicle is explicitly displaced from its cell $i$ to
\begin{equation}
j = \left(i + \hat{v}_t^{\text{disc}}(i)\right) \bmod L
    \label{eq:next_cell}
\end{equation}
and the next road state is constructed according to
\begin{equation}
    \hat{s}_t(j) = 
    \begin{cases}
        \hat{v}_t^{\text{disc}}(i), & \text{if cell } j \text{ is occupied by the vehicle moving from cell } i, \\
        -1, & \text{otherwise}.
    \end{cases}
\end{equation}

Since one updated vehicle is generated for every occupied cell in $\mathcal{C}_{t-1}$ rather than predicting occupancy independently, PI-NCA explicitly includes vehicle conservation into the learned dynamics, provided that the velocities are correctly computed.

\subsection{Stochastic PI-NCA Formulation}
\label{subsec:categorical_cca}

The deterministic PI-NCA assigns a single velocity to each vehicle. However, as seen in the background section (Subsection~\ref{subsec:CA_traffic_flow}) on the fundamental diagram, the flow is not directly a function of the local density. In terms of TCA, that means that the velocity $v_t(i)$ of the vehicle in cell $i$ is not a function of the distance $d_t(i)$ but rather stochastic (as proposed in the NaSch model). A solution would then be to consider mixture density networks \cite{bishop1994mixture} as a way to model a probability distribution over velocities with multiple modes. 
We therefore replace the scalar velocity output with a discrete probability distribution over the admissible velocities $0, \dots, v_{\max}$. The physical interpretation remains the same: the NCA still describes velocity locally, while the explicit update still determines vehicle movement.

For each occupied cell $i\in\mathcal{C}_{t-1}$, the output layer of the network produces $v_{\max} +1$ logits
\begin{equation}
    \mathbf{z}_t(i) = \left(z_{t,0}(i), z_{t,1}(i), \dots, z_{t,v_{\max}}(i)\right),
    \label{eq:logits}
\end{equation}
where each logit $z_{t,k}(i)$ is the unnormalized score the network assigns to velocity class $k\in\{0, \dots, v_{\max}\}$ for the vehicle occupying cell $i$ at time $t$. 
These logits are transformed into a categorical probability distribution through the Softmax function,
\begin{equation}
    P\left(v_t(i) = k \mid \mathbf{s}_{t-1}\right) = \frac{\exp\left(z_{t,k}(i)\right)}{\displaystyle\sum_{r=0}^{v_{\max}}\exp\left(z_{t,r}(i)\right)}, \qquad k=0,\dots,v_{\max}.
    \label{eq:categorical_softmax}
\end{equation}
% Thus, for each vehicle, the model represents the probabilities
% \begin{equation}
%     \left[P\left(v_t(i) = 0\right), P\left(v_t(i) = 1\right), \dots, P\left(v_t(i) = v_{\max}\right)\right],
% \end{equation}
The resulting probabilities are non-negative and sum to one over all admissible velocity classes. They therefore define the model's conditional distribution over each vehicle's possible next velocities.

Training is performed using categorical cross-entropy,
\begin{equation}
    \mathcal{L}_{\mathrm{cat}}(\theta) = -\frac{1}{|\mathcal{C}_{t-1}|}\sum_{i\in\mathcal{C}_{t-1}}\log P\left(v_t(i) \mid \mathbf{s}_{t-1}\right).
    \label{eq:categorical_loss}
\end{equation}
where $v_t(i)$ is the reference velocity at time $t$ of the vehicle occupying cell $i$ at time $t-1$. Thus, the loss maximizes the probability the model assigns to the observed reference velocity class.

During inference, a velocity $\tilde{v}_t(i)$ is sampled from the learned conditional distribution,
\begin{equation}
    \tilde{v}_t(i) \sim \mathrm{Categorical}\left(P\left(v_t(i) = 0\mid \mathbf{s}_{t-1}\right), \dots, P\left(v_t(i) = v_{\max}\mid \mathbf{s}_{t-1}\right)\right).
\end{equation}
We then use the sampled velocity in the same explicit update mechanism as in the deterministic PI-NCA. A vehicle occupying cell $i$ at time $t-1$ is displace to
\begin{equation}
j = \left(i + \tilde{v}_t(i)\right) \bmod L
    \label{eq:next_cell_stoch}
\end{equation}
with the next road state being constructed according to
\begin{equation}
    \tilde{s}_t(j) = 
    \begin{cases}
        \tilde{v}_t(i), & \text{if cell } j \text{ is occupied by the vehicle moving from cell } i, \\
        -1, & \text{otherwise}.
    \end{cases}
\end{equation}
Consequently, repeated rollouts from the same initial traffic state may generate different trajectories, while the local representation and vehicle conservation structure of the deterministic PI-NCA are preserved.

\begin{figure}[ht]
    \centering
    \includegraphics[width=0.8\linewidth]{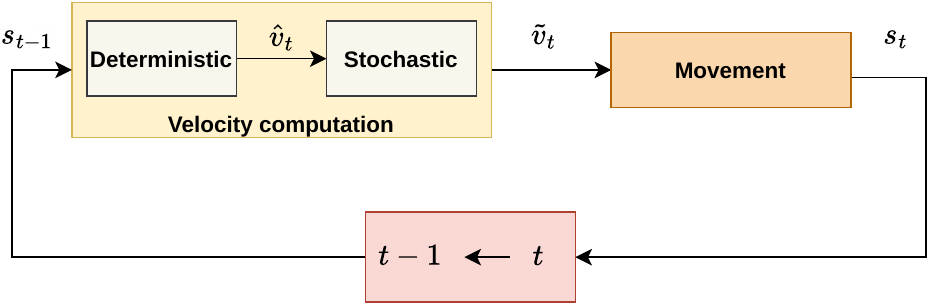}
    \caption{Conceptual progression from deterministic to stochastic PI-NCA. The stochastic formulation extends the deterministic velocity prediction to a categorical velocity distribution while retaining the explicit movement update.}
    \label{fig:nca_diagram}
\end{figure}

Figure~\ref{fig:nca_diagram} summarizes the conceptual progression from the deterministic to the stochastic PI-NCA formulation. The deterministic PI-NCA first introduces the decomposition between velocity prediction and explicit vehicle movement. The stochastic PI-NCA builds on this formulation by extending the velocity prediction from a single deterministic value $\hat{v}_t$ to a sampled velocity $\tilde{v}_t$ obtained from a learned categorical distribution, while retaining the same explicit movement mechanism. 

\section{Training and Evaluation}
\label{subsec:training_evaluation}

We discuss here the practical training and evaluation details in two simulation cases.

\subsection{NaSch experiments}

We use a maximum velocity of $v_{\text{max}}=5$ and a kernel of size $k=11$ for the first convolution. Therefore, the local prediction has access to five cells on either side of the cell under consideration. The other layers use a kernel size of 1 and 16 hidden channels, as shown in Figure~\ref{fig:cnn_architecture_nca}.

\begin{figure}[ht]
    \centering
    \includegraphics[width=1.0\linewidth]{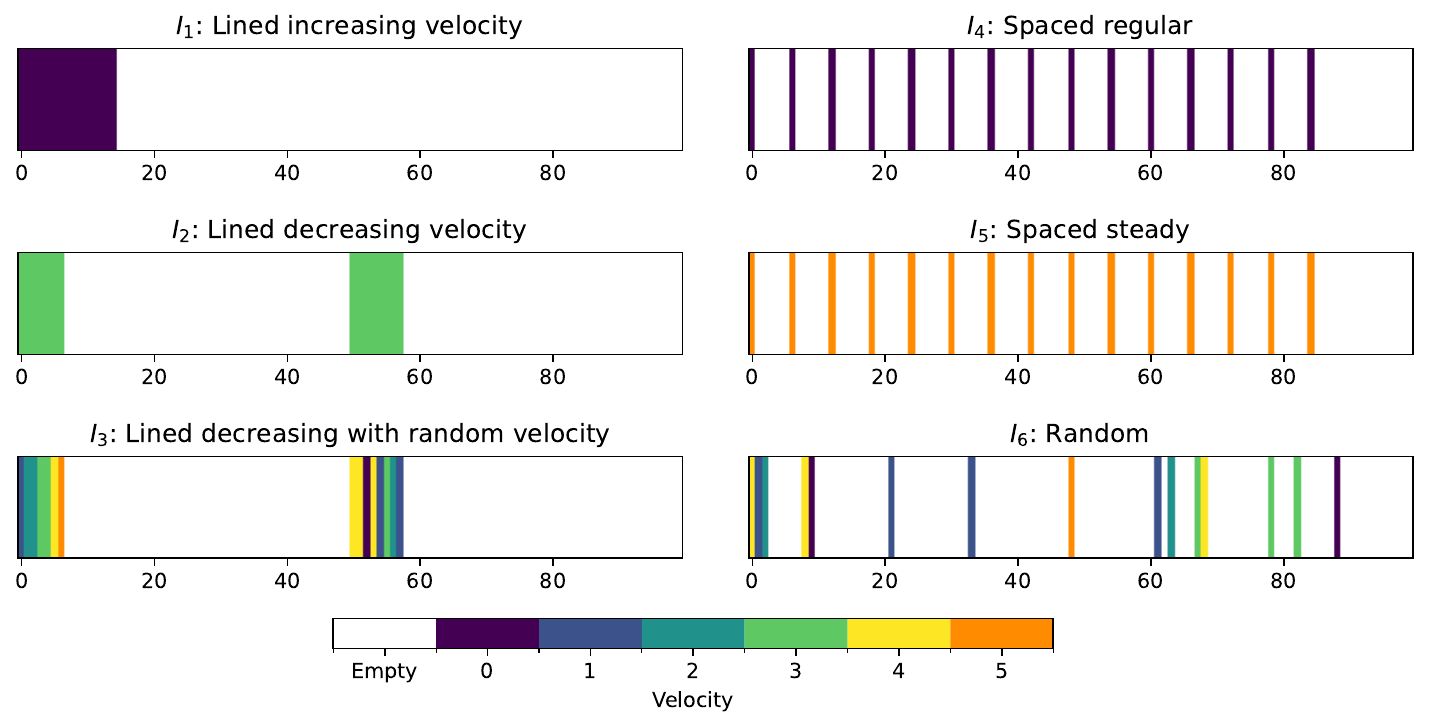}
    \caption{Six initial states used for training and testing.}
    \label{fig:initial_states}
\end{figure}

Training and test data consist of trajectories generated using the NaSch reference model on a single-lane road with periodic boundary conditions. The training dataset covers different vehicle densities and spatial traffic configurations. It includes six initialization schemes, as shown in Figure \ref{fig:initial_states}, ranging from structured to randomly generated configurations. For simplicity, we call them $I_1-I_6$: $I_1$, lined increasing velocity, has lined vehicles next to each others starting with velocity zero; $I_2$, lined decreasing velocity, is similar to $I_1$ but vehicles start at higher velocities, and there are two blocks of cars; $I_3$, lined decreasing with random velocity, has also two blocks of cars adjacent to each others, but their velocity in this case is random; $I_4$, spaced regular, has vehicles regularly spaced starting with velocity zero; $I_5$, space steady, has also spaced vehicles but starting with maximum velocity; $I_6$, random, has vehicles placed randomly on the road, with each vehicle assigned a random initial velocity from $\{0, \dots, v_{\max}\}$. For each vehicle count, we generate 350 trajectories: 50 for each structured initialization and 100 for the random initialization, providing a wider variety of local traffic configurations. 

We generate the test dataset independently of the training set using unseen random seeds. We evaluate on similar road configurations with the same initial states as for the training conditions, but we also include vehicle counts and road lengths not encountered before. This tests how well the model generalizes to unseen cases. For each test configuration, we consider 10 independently initialized trajectories. Table \ref{tab:training_testing_parameters} reports a summary of the main corresponding training and testing conditions.

\begin{table}[ht]
\centering
\caption{Data-generation settings using the NaSch model for training and testing.}
\label{tab:training_testing_parameters}

\small
\setlength{\tabcolsep}{15pt}
\renewcommand{\arraystretch}{1.25}

\begin{tabular}{@{}lccc@{}}
\toprule
\textbf{Parameter}
& \textbf{Training}
& \textbf{Testing} \\
\midrule

Road length $L$
& 100
& 50, 100, 150 \\

Number of vehicles $N$
& 10, 15, 20
& 8, 10, 12, 15, 17, 20 \\

Maximm velocity $v_{\text{max}}$
& 5
& 5 \\

Trajectory length $T$
& 40
& 40 \\

Initialization types
& 6
& 6 \\

\bottomrule
\end{tabular}
\end{table}

We run the NCA and PI-NCA independently five times with different random weight initializations. The architecture and optimization parameters are fixed across runs and summarized in Table~\ref{tab:hyperparameters_models}.

\begin{table}[ht]
\centering
\caption{NCA and PI-NCA training configuration for deterministic and stochastic NaSch experiments.}
\label{tab:hyperparameters_models}

\small
\setlength{\tabcolsep}{15pt}
\renewcommand{\arraystretch}{1.25}

\begin{tabular}{@{}lccc@{}}
\toprule
\textbf{Hyperparameter}
& \textbf{Value} \\
\midrule

Optimizer
& Adam \\

Learning rate
& $10^{-3}$ \\

Weight decay
& $10^{-5}$ \\

Batch size
& 64 \\

Hidden channels
& 16 \\

Kernel size
& 11 \\

Deterministic training epochs
& 3000 \\

Stochastic training epochs
& 1000 \\

\bottomrule
\end{tabular}
\end{table}

For the stochastic experiments, we train the stochastic PI-NCA on trajectories generated with $p>0$; in particular, we train separate models for each $p$ from 0.1 to 0.9. Each model therefore learns the transition distribution associated with a specific level of stochasticity in the NaSch dynamics. As in the deterministic cases, we perform five training runs with different random seeds for each value of $p$.
We evaluate all models using autoregressive rollout, meaning predicted traffic states are recursively used as inputs for next-step predictions.

Prediction performance is assessed using the Mean Absolute Error (MAE), Mean Squared Error (MSE), Car Mean Absolute Error (Car MAE), Occupancy Accuracy (OA), Conservation Error (CE), and Relative Conservation Error (RCE). For stochastic PI-NCA, we also report the Total Variation (TV) distance compared to the corresponding NaSch model. It directly measures the agreement between the next-velocity probability distributions predicted by PI-NCA and those obtained by the reference NaSch model. Finally, we evaluate relative $L^2$ errors to compare global density--flow relations between PI-NCA and the reference models. We provide their formal definitions in~\ref{appendix:metrics}.

\subsection{Extension to KKW-1 Dynamics}
\label{subsec:extension_kkw1}

To assess the overall generalization of the proposed NCA and PI-NCA formulations, we additionally apply them to learn the dynamics of the KKW-1 model, a cellular automaton based on the three-phase traffic theory of Kerner, Klenov, and Wolf \cite{kerner2002cellular}. In the NaSch model, vehicles interact primarily through the available gap, while KKW-1 introduces a synchronization distance that helps vehicles adapt their velocity to the car ahead. Therefore, the transition rules depend on the available gap between cars and the leading vehicle's velocity; as a result, the local interactions are more complex to study.

\begin{figure}[ht]
     \centering
     \begin{subfigure}[b]{0.49\textwidth}
         \centering
         \includegraphics[width=\textwidth]{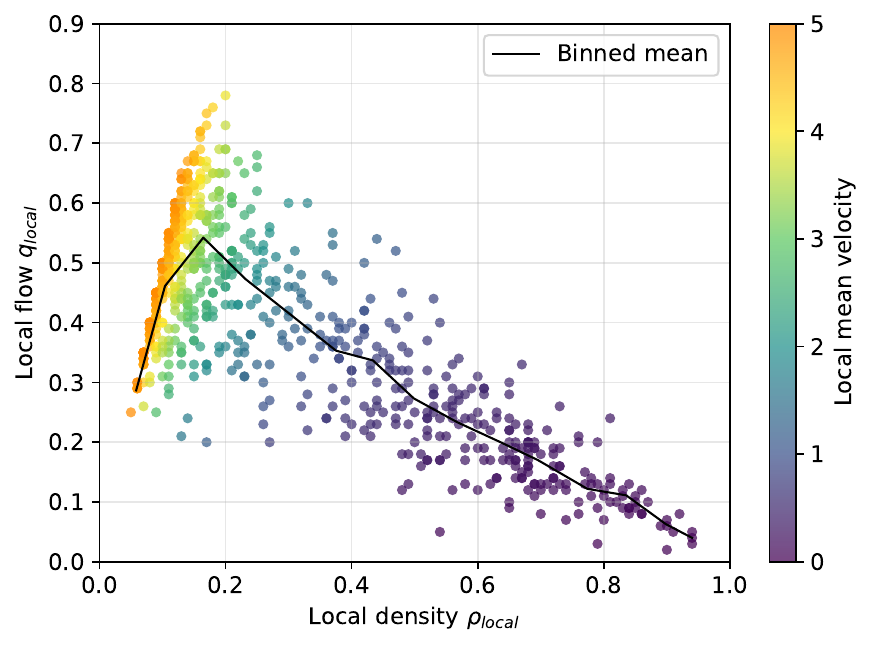}
         \caption{Local fundamental diagram}
         \label{fig:local_fd_kkw1}
     \end{subfigure}
     \hfill
     \begin{subfigure}[b]{0.49\textwidth}
         \centering
         \includegraphics[width=\textwidth]{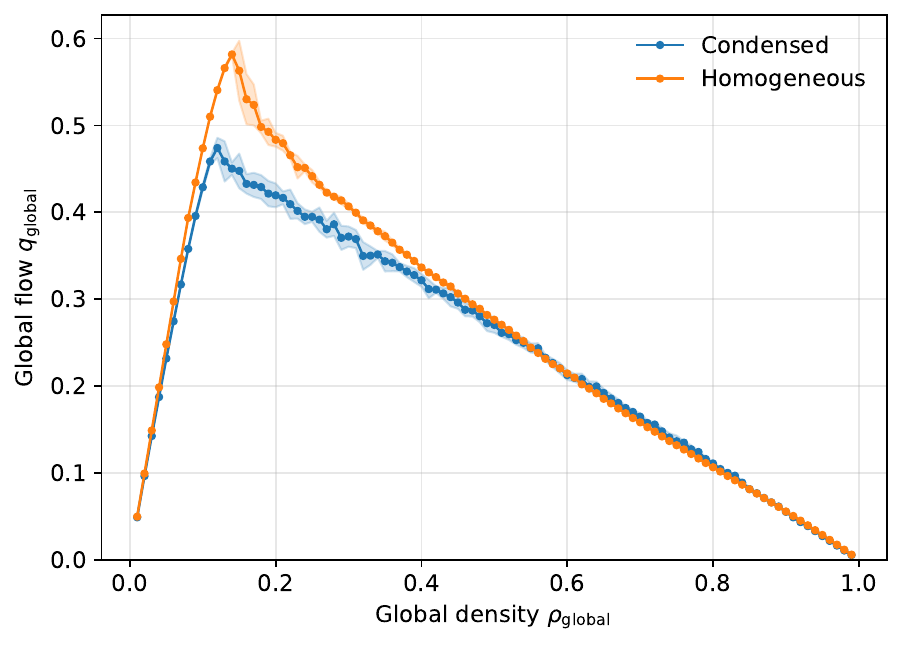}
         \caption{Global fundamental diagram}
         \label{fig:global_fd_kkw1}
     \end{subfigure}
\caption{Fundamental diagrams for KKW-1 model.}
\label{fig:fds_kkw1}
\end{figure}

The KKW-1 dynamics are adapted to the discrete road representation used for the NaSch model, with a maximum velocity of $v_{\text{max}}=5$ and periodic boundary conditions. We rescale the synchronization distance and stochastic parameters accordingly. \ref{appendix:kkw1_reference_dynamics} provides the full update equations and parameterization. 
Figure~\ref{fig:fds_kkw1} illustrates the resulting local and global traffic behavior of the KKW-1 model. 
The local fundamental diagram shows a broad distribution of traffic states, highlighting spatial and temporal variability in local traffic dynamics. The corresponding binned mean shows the transition from increasing flow at low densities to decreasing flow as congestion evolves. The plot shows one simulation with 5000 iterations, $L=100$, and parameters as given in Table~\ref{tab:kkw1_parameters}. The global fundamental diagram complements the local representation, showing the relation between density and flow for the whole system. In particular, the condensed and homogeneous initial configurations produce distinct flows over intermediate-density ranges, then converge at higher densities. Therefore, in this range, the macroscopic traffic behavior depends on the initial traffic configuration. For this experiment, we consider a periodic road of $L=1000$ cells, and for each density, we consider both condensed and homogeneous initial traffic configurations. We simulate each realization for 1000 time steps and average over 5 independent stochastic realizations.

Because the KKW-1 synchronization rule can depend on vehicles farther ahead than the NaSch rule, it requires a larger spatial receptive field. Therefore, we increase the main convolution kernel from 11 to 23 and introduce an additional spatial convolution with a kernel size of 10 to further enlarge the effective receptive field. Moreover, the training and evaluation remain unchanged except that we use 32 hidden channels instead of 16.

%%%%%%%%%%%%%%%%%%%%%%%%%%%%%%%%%%%%%%%%%%%%%%%%%%%%%%%%%%%%%%%%%%
\section{Results}
	\label{sec:results}

\subsection{Deterministic Prediction Performance}

We first evaluate NCA and PI-NCA's ability to reproduce the deterministic dynamics of the reference traffic CA. Table~\ref{tab:global} summarizes the prediction performance over the complete NaSch and KKW-1 test sets. The results report the mean and standard deviation over five independently trained models.

\begin{table}[ht]
\centering
\caption{Overall prediction performance over the complete NaSch and KKW-1 test sets.}
\label{tab:global}

\small
\setlength{\tabcolsep}{4pt}
\renewcommand{\arraystretch}{1.25}

\begin{tabular}{@{}llccc@{}}
\toprule
\textbf{Traffic CA}
& \textbf{Model}
& \textbf{MAE}
& \textbf{Car MAE}
& \textbf{CE} \\
\midrule

\multirow{2}{*}{NaSch}
& NCA
& $0.0022 \pm 0.0015$
& $0.0070 \pm 0.0062$
& $0.0071 \pm 0.0037$ \\

& PI-NCA
& $0.00006 \pm 0.00011$
& $0.00009 \pm 0.00015$
& $0.00018 \pm 0.00040$ \\

\midrule

\multirow{2}{*}{KKW-1}
& NCA
& $0.1949 \pm 0.0289$
& $0.4667 \pm 0.0768$
& $0.3085 \pm 0.0839$ \\

& PI-NCA
& $0.0179 \pm 0.0090$
& $0.0391 \pm 0.0187$
& $0.00017 \pm 0.00038$ \\

\bottomrule
\end{tabular}
\end{table}

For NaSch, both approaches achieve high prediction accuracy, with the PI-NCA consistently outperforming the NCA across the reported metrics. The improvement is evident in both overall state prediction and vehicle-specific error. The lower conservation error for PI-NCA indicates that the explicit update rule more reliably preserves the number of vehicles.
PI-NCA's advantage is even more pronounced under the more complex dynamics of the KKW-1 model. While NCA shows larger state and vehicle errors than for NaSch, the PI-NCA maintains comparatively low prediction errors and near-zero conservation error. This result suggests that predicting vehicle velocities and explicitly updating their positions provide an effective inductive bias for learning more complex dynamics. We report an extension of these results in~\ref{appendix:additional_performance_tables}.

Figure~\ref{fig:deterministic_simulation} illustrates representative NaSch rollout behavior when NCA cannot conserve the exact number of cars in the simulation. 
The error arises at the second time step, where an extra vehicle randomly appears, propagating the error forward in time. 
 
\begin{figure}[ht]
    \centering
    \includegraphics[width=1.0\linewidth]{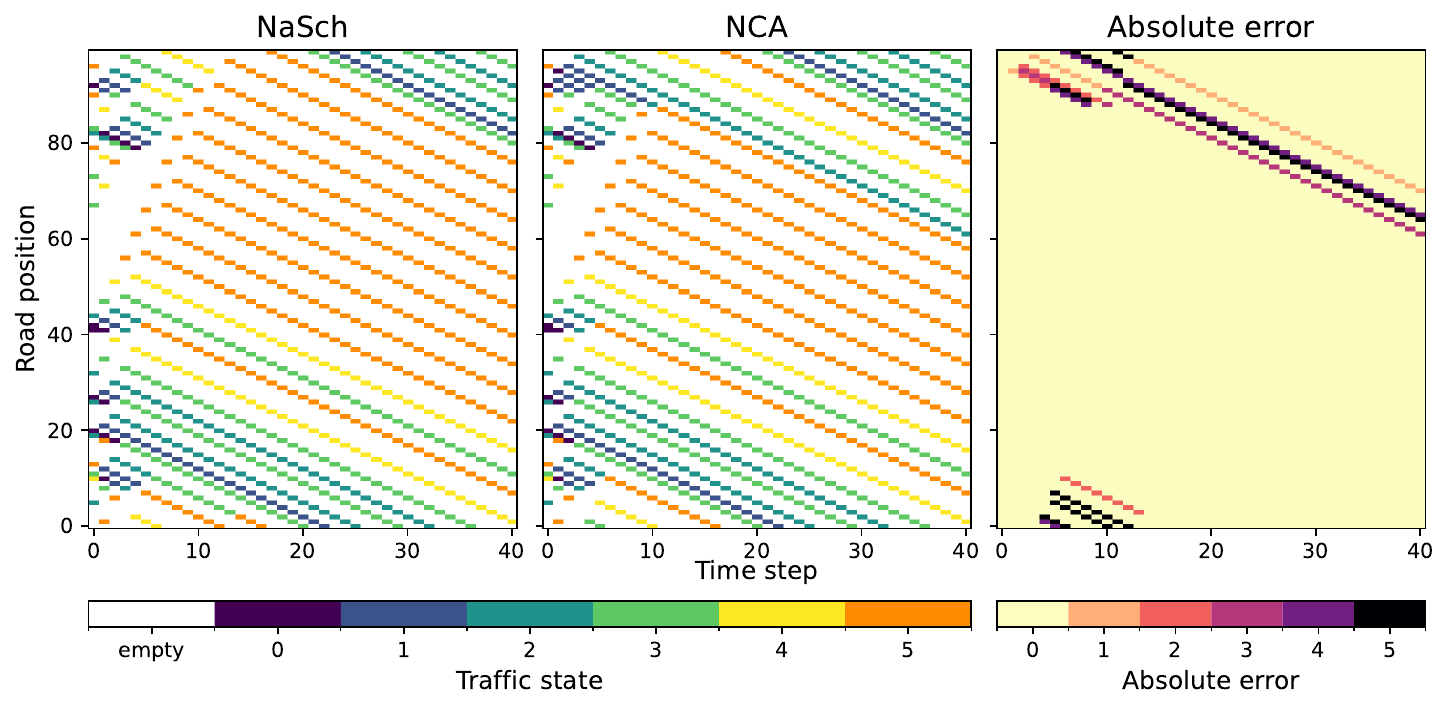}
    \caption{Comparison of NaSch (left) with the proposed NCA (middle) and their absolute error (right), using $L=100$, 20 vehicles, $T=40$, and random initialization.}
    \label{fig:deterministic_simulation}
\end{figure}

To evaluate the robustness of the proposed models, we assess their generalization across traffic initializations, road lengths, and vehicle densities not necessarily observed during training. In all plots, each point represents the performance of an independently trained model under the specific test condition, highlighting variability across runs.

\subsubsection{Generalization Across Initial Conditions}

\begin{figure}[h!]
     \centering
     \begin{subfigure}[b]{1.0\textwidth}
         \centering
         \includegraphics[width=\textwidth]{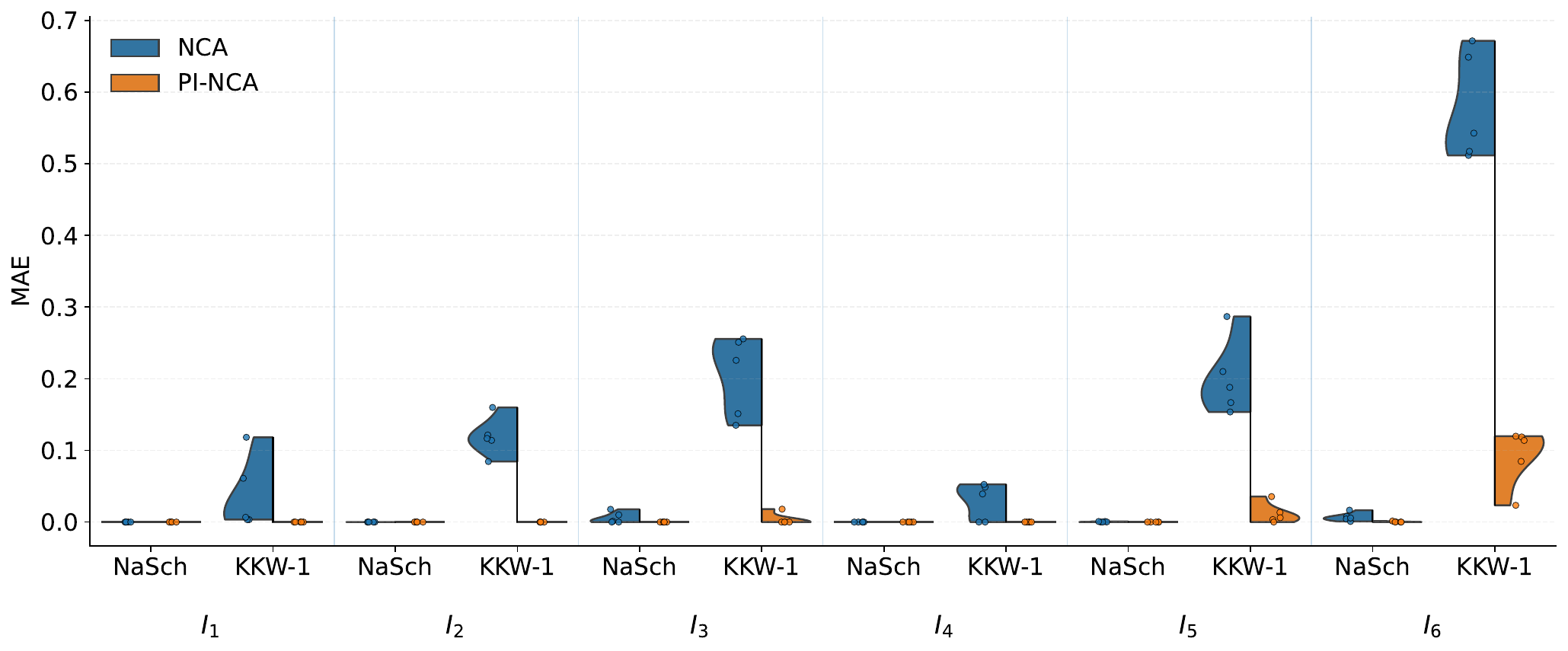}
         \caption{MAE}
         \label{fig:mae_initcase}
     \end{subfigure}
     \hfill
     \begin{subfigure}[b]{1.0\textwidth}
         \centering
         \includegraphics[width=\textwidth]{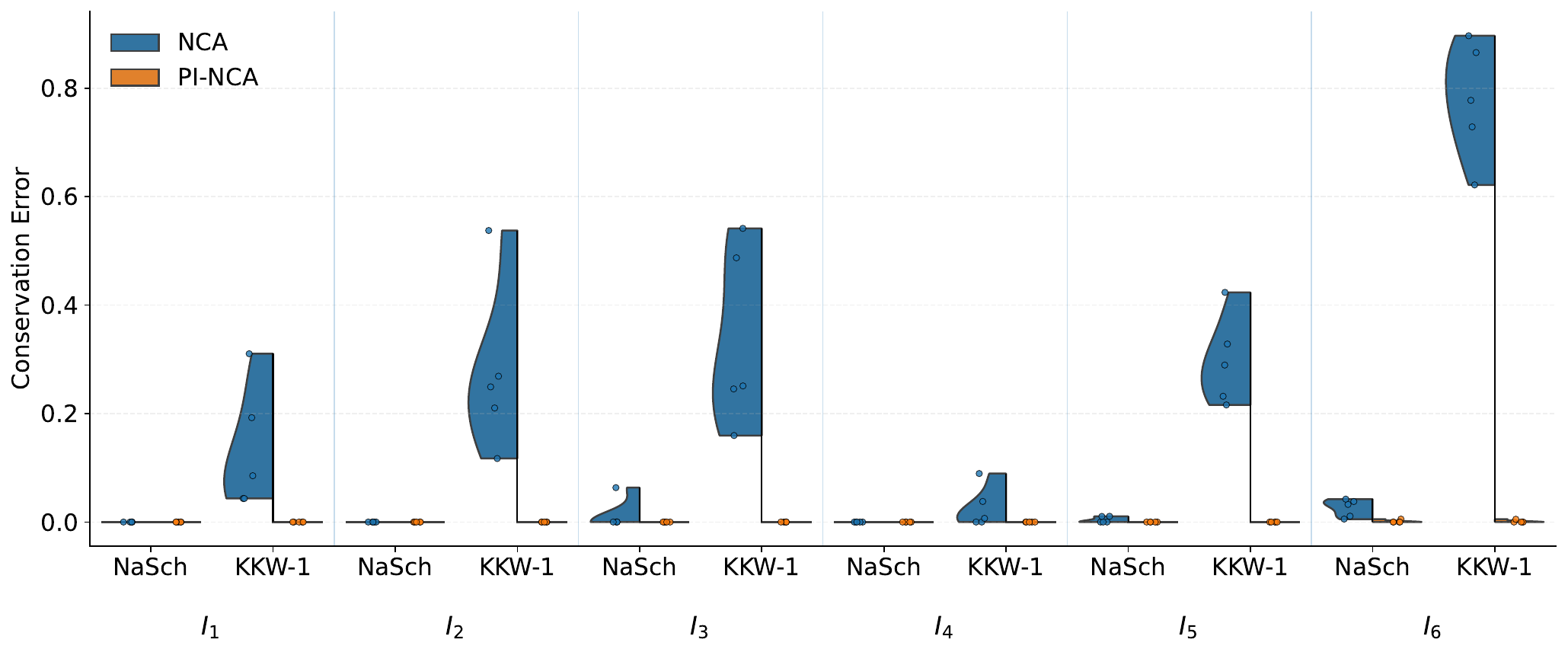}
         \caption{Conservation Error}
         \label{fig:ce_initcase}
     \end{subfigure}
\caption{Generalization across traffic initializations for the deterministic NCA (blue) and PI-NCA (orange) with NaSch and KKW-1 dynamics. (a) Mean absolute error and (b) conservation error. Each point corresponds to an independent training run.}
\label{fig:results_initcase}
\end{figure}

Figure~\ref{fig:results_initcase} compares the models across the different traffic initializations. For NaSch, both NCA and PI-NCA maintain small errors for all the tested initial conditions. By contrast, KKW-1 shows a notable difference between the two proposed models. NCA error depends on initialization and is larger for random configurations, whereas the physics-informed counterpart remains more consistent and accurate across all initial conditions. Similarly, for vehicle conservation, KKW-1 NCA shows the largest errors, particularly for the random configurations. In contrast, PI-NCA conservation error remains close to zero for all tested initializations. 
Table \ref{tab:initcase} reports the specifics of the prediction errors.

\subsubsection{Generalization Across Road Lengths}

\begin{figure}[h!]
     \centering
     \begin{subfigure}[b]{1.0\textwidth}
         \centering
         \includegraphics[width=\textwidth]{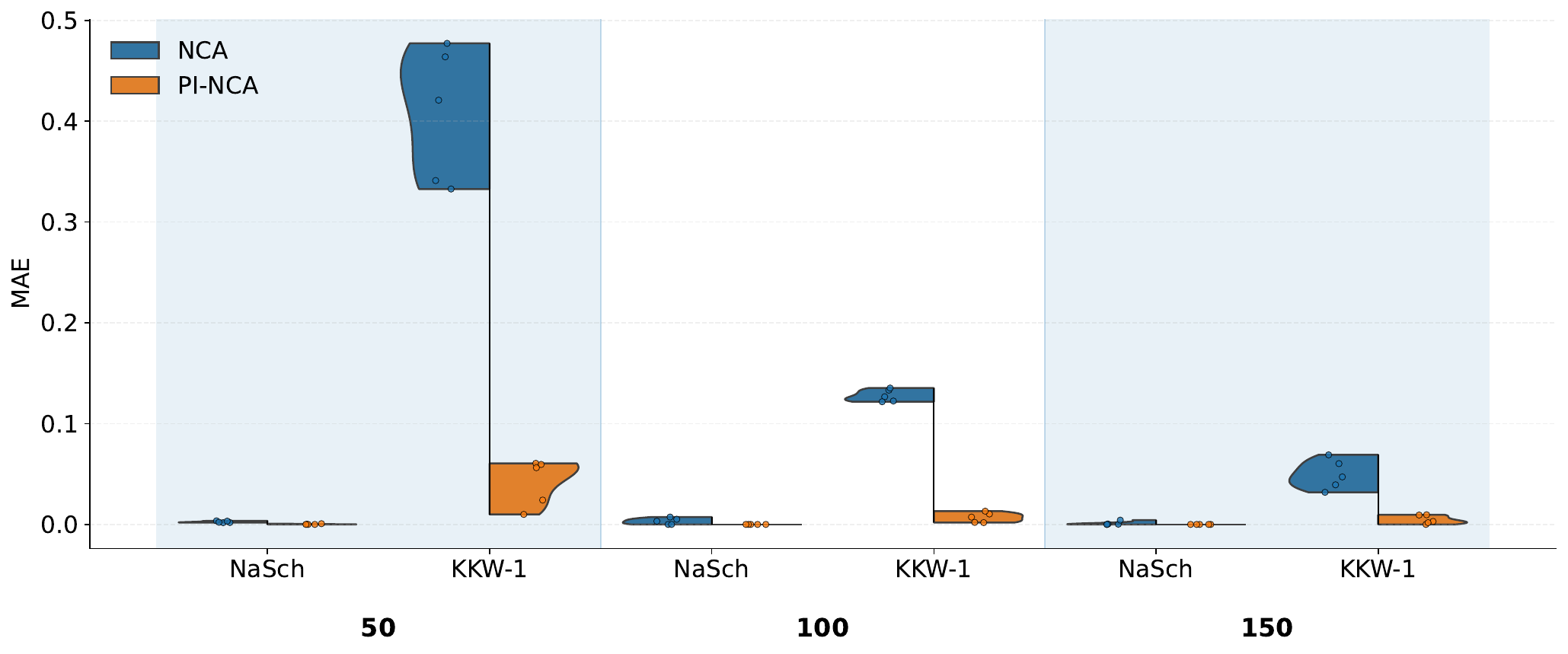}
         \caption{MAE}
         \label{fig:mae_L}
     \end{subfigure}
     \hfill
     \begin{subfigure}[b]{1.0\textwidth}
         \centering
         \includegraphics[width=\textwidth]{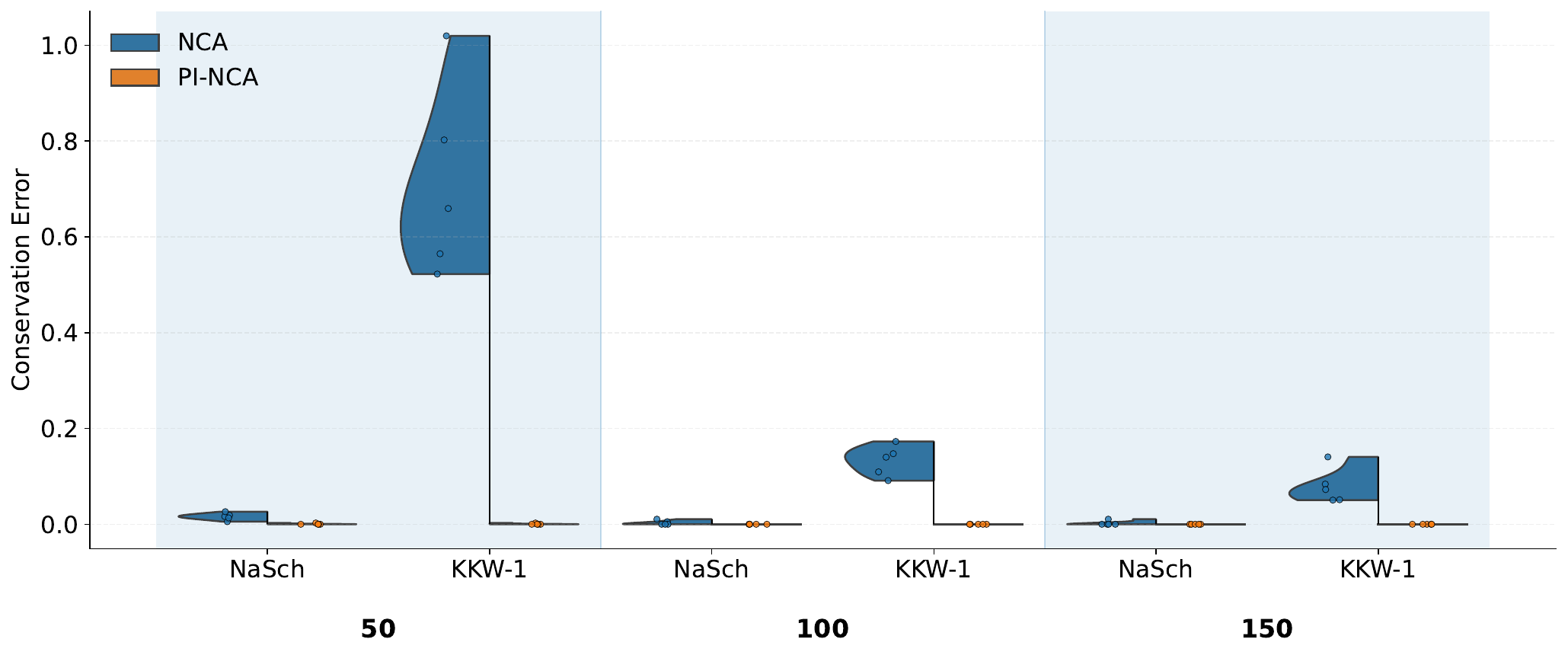}
         \caption{Conservation Error}
         \label{fig:ce_L}
     \end{subfigure}
\caption{Generalization across road lengths for the deterministic NCA (blue) and PI-NCA (orange) with NaSch and KKW-1 dynamics. (a) Mean absolute error and (b) conservation error for $L\in \{50, 100, 150\}$. Shaded regions indicate road lengths not used during training. Each point corresponds to an independent training run.}
\label{fig:results_L}
\end{figure}

Figure~\ref{fig:results_L} evaluates the models on road lengths both inside and outside the training distribution. For NaSch, prediction errors remain low across all three road lengths. For KKW-1, the PI-NCA produces lower MAE than the NCA, even for the unseen road lengths $L=50$ and $L=150$. The physics-informed model also conserves cars better. The KKW-1 NCA shows high conservation error, especially for $L=50$, while the PI-NCA models achieve near-zero error across all road lengths.
Table~\ref{tab:L} reports the errors.

\subsubsection{Generalization Across Vehicle Numbers}

\begin{figure}[h!]
     \centering
     \begin{subfigure}[b]{1.0\textwidth}
         \centering
         \includegraphics[width=\textwidth]{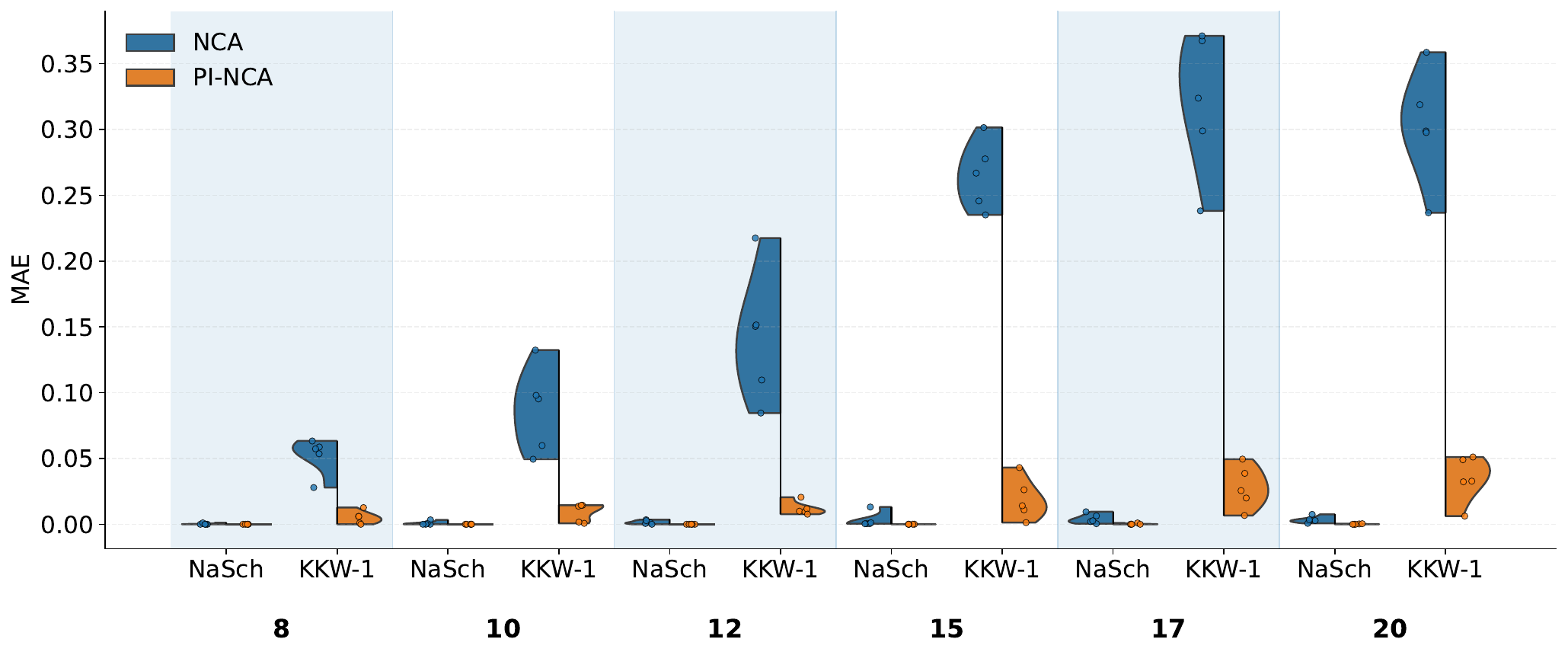}
         \caption{MAE}
         \label{fig:num_cars_mae}
     \end{subfigure}
     \hfill
     \begin{subfigure}[b]{1.0\textwidth}
         \centering
         \includegraphics[width=\textwidth]{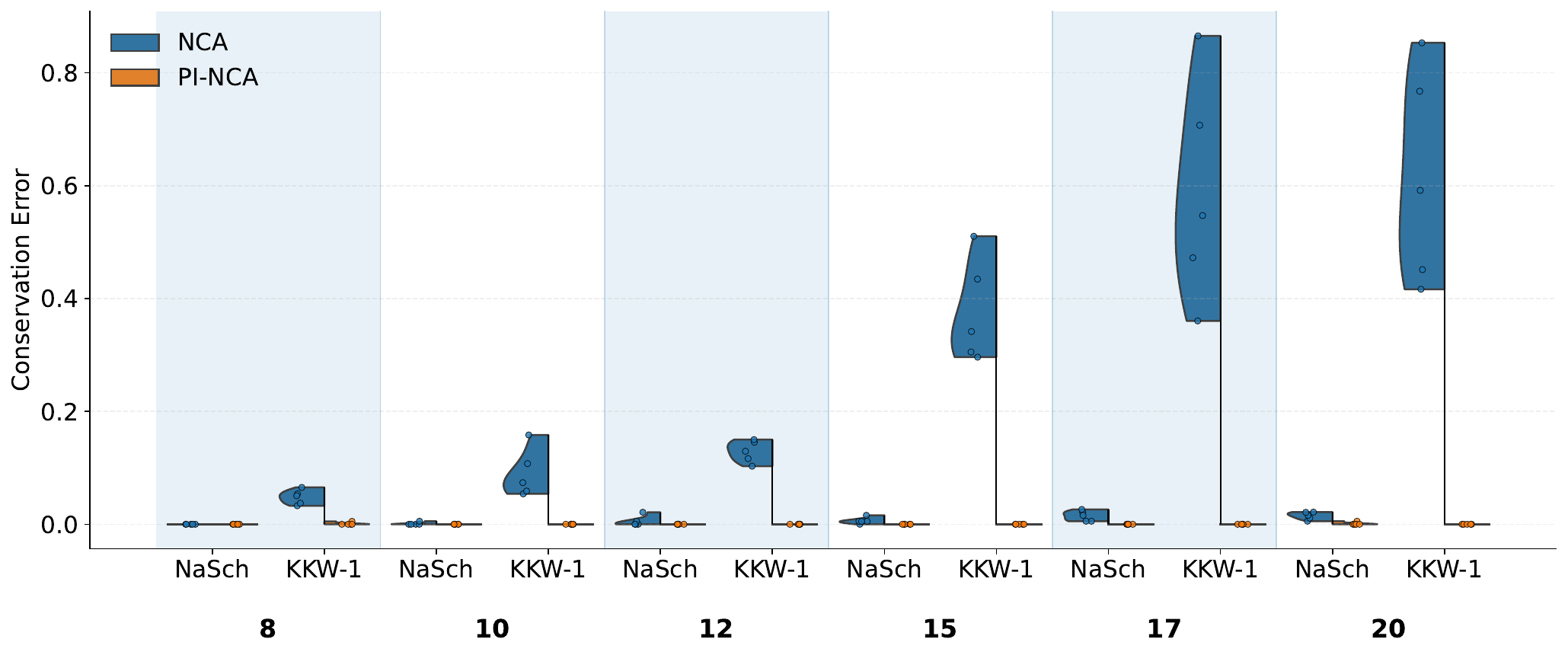}
         \caption{Conservation Error}
         \label{fig:num_cars_ce}
     \end{subfigure}
\caption{Generalization across vehicle numbers for the deterministic NCA (blue) and PI-NCA (orange) with NaSch and KKW-1 dynamics. (a) Mean absolute error and (b) conservation error. Shaded regions indicate vehicle numbers not used during training. Each point corresponds to an independent training run.}
\label{fig:results_num_cars}
\end{figure}

Finally, Figure~\ref{fig:results_num_cars} compares prediction performance on the number of vehicles observed during training with unseen densities. The NaSch model remains highly accurate across all vehicle counts tested. For KKW-1 dynamics, the NCA errors, both MAE and conservation error, increase as the number of vehicles increases. PI-NCA instead keeps MAEs low for both seen and unseen numbers of vehicles, with near-zero conservation errors.
Table~\ref{tab:density} reports the errors.

Results across different initializations, road lengths, and vehicle counts show that the proposed physics-informed model improves robustness when the learned rules are applied outside the training configurations.

\subsection{Stochastic Prediction Performance}
\label{subsec:stochastic_pred_performance}

In the deterministic experiments, the proposed models aim to replicate the reference models' transition rules. For stochastic dynamics, we cannot evaluate a single state evolution in the same way, as the same initial state can produce different traffic trajectories. Therefore, we assess the stochastic PI-NCA in three ways: the resulting traffic evolution in space and time, the emergent local traffic behavior, and the learned conditional velocity distributions.

\subsubsection{Stochastic Traffic Evolution}

\begin{figure}[h!]
     \centering
     \begin{subfigure}[b]{0.8\textwidth}
         \centering
         \includegraphics[width=\textwidth]{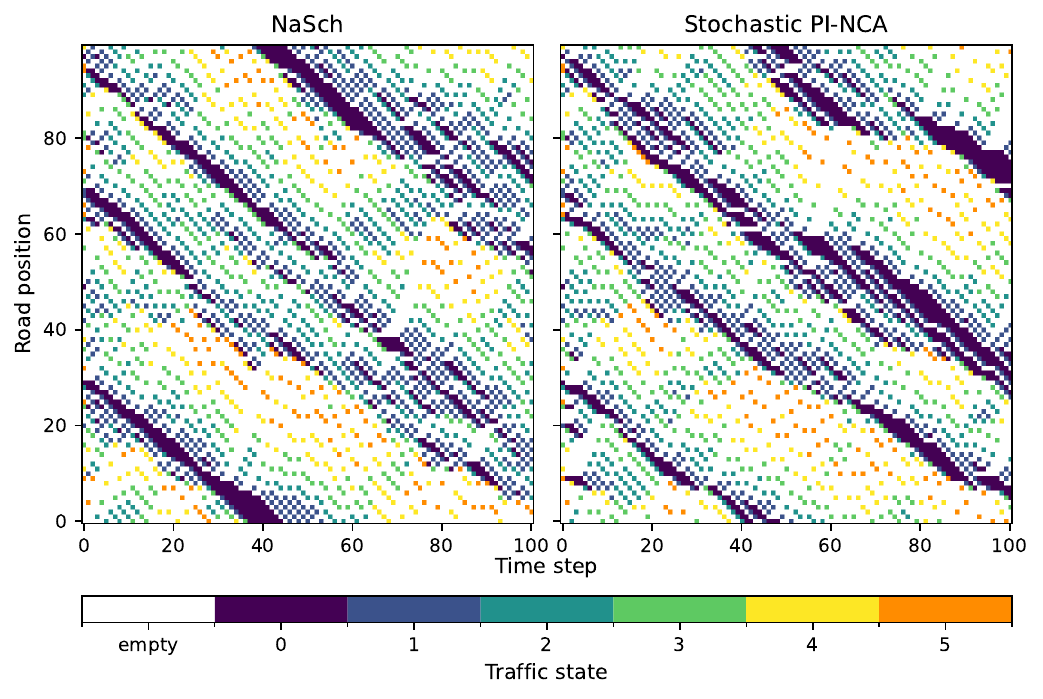}
         \caption{NaSch dynamics}
         \label{fig:stochastic_nasch_pinca}
     \end{subfigure}
     \hfill
     \begin{subfigure}[b]{0.8\textwidth}
         \centering
         \includegraphics[width=\textwidth]{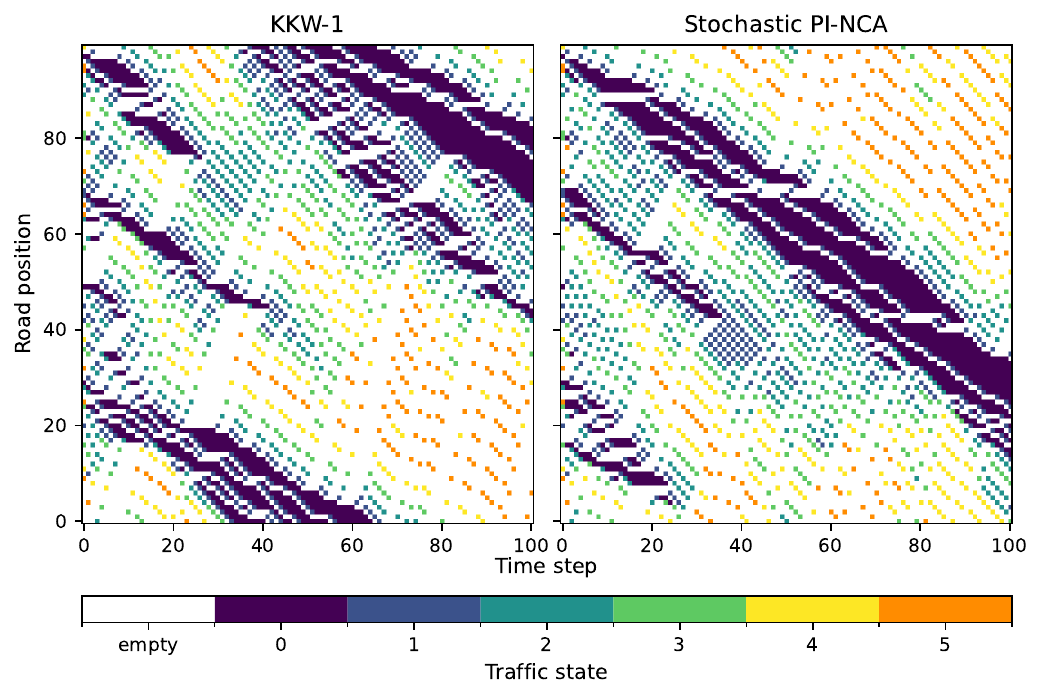}
         \caption{KKW-1 dynamics}
         \label{fig:stochastic_kkw1_pinca}
     \end{subfigure}
\caption{Representative stochastic space-time trajectories generated by the reference traffic CA and the stochastic PI-NCA. (a) NaSch with $p=0.20$, $L=100$, $\rho=0.30$, $T=100$, and random initialization; (b) KKW-1 using the stochastic parameters in Table \ref{tab:kkw1_parameters} with $L=100$, $\rho=0.30$, $T=100$, and random initialization.}
\label{fig:stochastic_pinca}
\end{figure}

Figure~\ref{fig:stochastic_pinca} shows representative stochastic rollouts from the reference models, NaSch and KKW-1, and stochastic PI-NCA. For NaSch, both simulations show characteristics of both freely moving vehicles and areas of slower, denser traffic. 
The figure also shows a good fit to the characteristic wave speed, even though the state itself is quite different. This demonstrates the learned model's ability to estimate local properties.

Similarly, for KKW-1, we notice the formation and propagation of lower-velocity regions in both the reference and learned dynamics. These examples provide a qualitative comparison to show whether the learned rule generates traffic evolution with spatial and temporal characteristics similar to those of the reference counterparts.

\subsubsection{Fundamental Diagram Analysis}
\label{subsubsec:fundamental_diagram_analysis}

\begin{figure}[h!]
     \centering
     \begin{subfigure}[b]{0.48\textwidth}
         \centering
         \includegraphics[width=\textwidth]{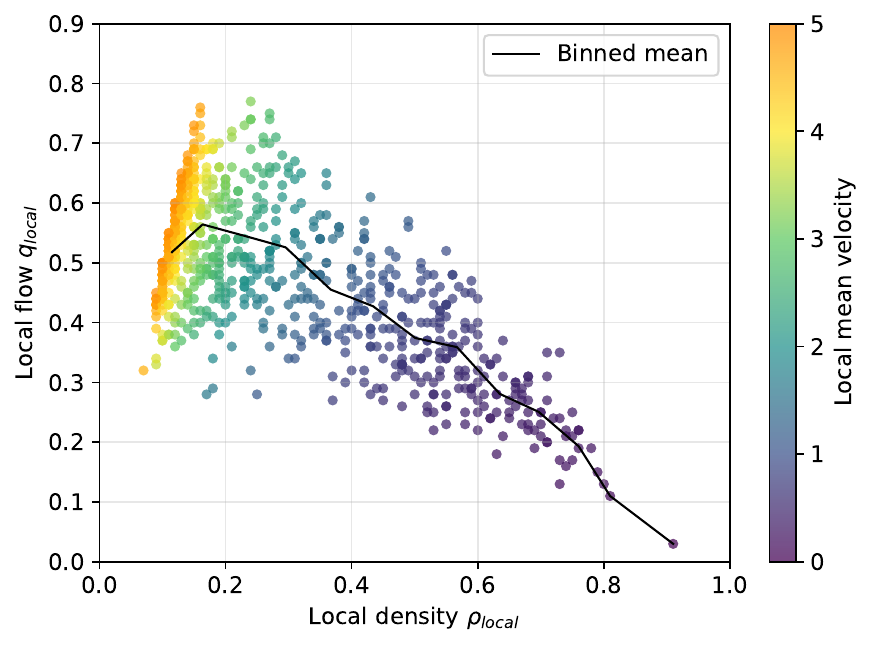}
         \caption{Local NaSch dynamics}
         \label{fig:local_fd_pinca_nasch}
     \end{subfigure}
     %\hfill
     \begin{subfigure}[b]{0.48\textwidth}
         \centering
         \includegraphics[width=\textwidth]{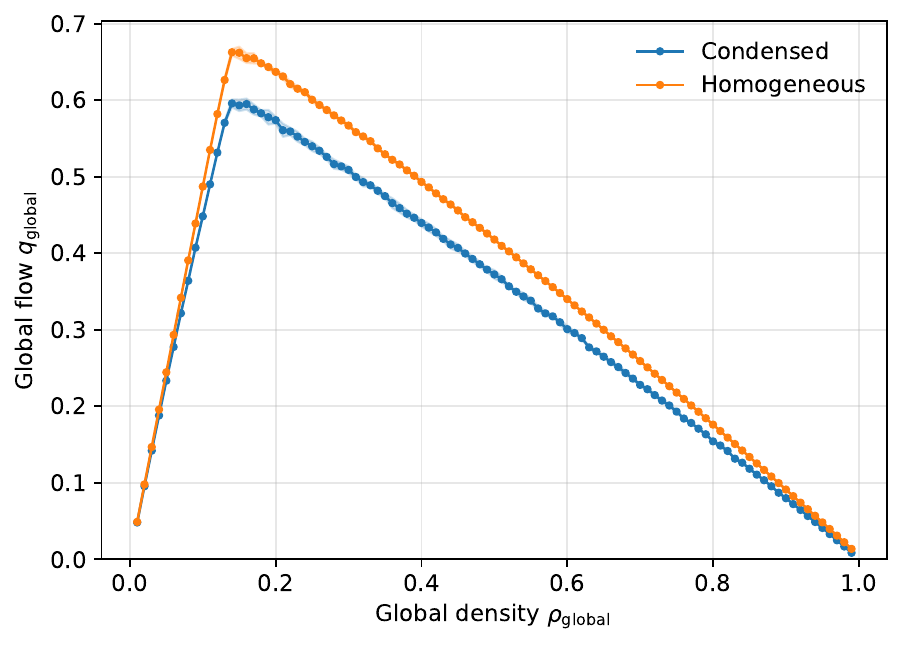}
         \caption{Global NaSch dynamics}
         \label{fig:global_fd_pinca_nasch}
     \end{subfigure}
     %\hfill
     \begin{subfigure}[b]{0.48\textwidth}
         \centering
         \includegraphics[width=\textwidth]{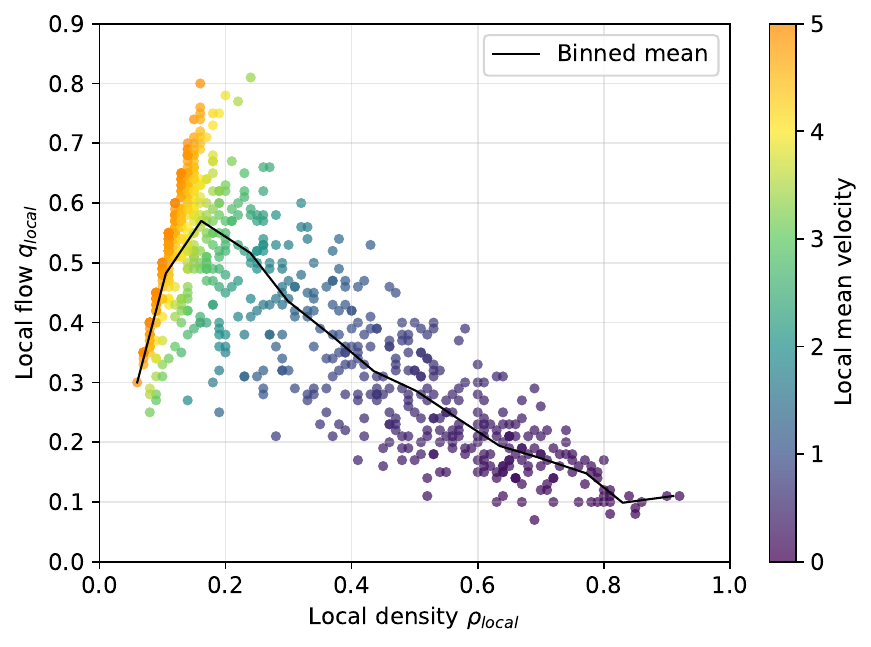}
         \caption{Local KKW-1 dynamics}
         \label{fig:local_fd_pinca_kkw1}
     \end{subfigure}
     %\hfill
     \begin{subfigure}[b]{0.48\textwidth}
         \centering
         \includegraphics[width=\textwidth]{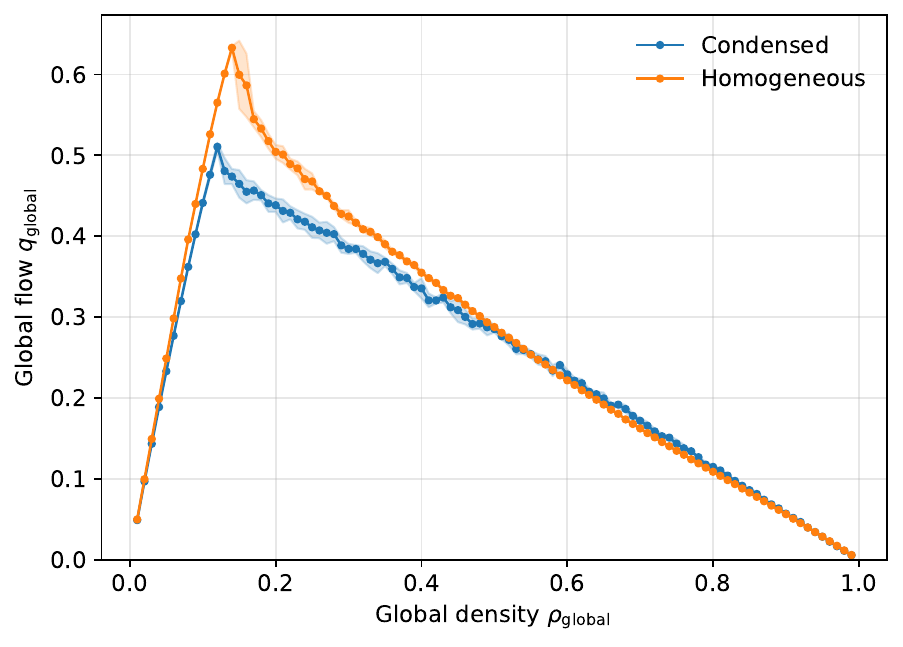}
         \caption{Global KKW-1 dynamics}
         \label{fig:global_fd_pinca_kkw1}
     \end{subfigure}
\caption{Local traffic behavior generated by the stochastic PI-NCA for NaSch and KKW-1 dynamics. (a)-(c) Local fundamental diagrams for NaSch and KKW-1, respectively. (b)-(d) Global fundamental diagrams for NaSch and KKW-1, respectively.}
\label{fig:stochastic_pinca_fds}
\end{figure}

To evaluate the traffic behavior emerging from longer autoregressive rollouts, we compare the local and global fundamental diagrams generated by the stochastic PI-NCA with those of the corresponding reference models presented earlier in Figures~\ref{fig:fds_nasch} and~\ref{fig:fds_kkw1}. Figure~\ref{fig:stochastic_pinca_fds} shows the PI-NCA results.

The local fundamental diagrams are obtained from a single stochastic simulation with $T=5000$ time steps. We obtain local traffic measurements following \cite{ziabari2026labeled} and using virtual spatial detectors over fixed temporal windows along the periodic road. For a road of length $L=100$, the detectors start at cells 10, 40, and 70, and each covers five consecutive cells. For each detector, we compute local density, flow, and mean velocity over consecutive 20-time-step windows. Thus, each point represents the traffic conditions observed during a single detector-window measurement. The solid line represents the binned mean local flow as a function of local density.

Figures~\ref{fig:local_fd_pinca_nasch} and \ref{fig:local_fd_pinca_kkw1} show that the stochastic PI-NCA reproduces the characteristic local density-flow structure of both reference models. At low local densities, traffic is characterized mostly by high velocities and increasing flow, whereas when congestion arises, lower mean velocities and decreasing flow appear. More importantly, multi-class categorical distributions allow the coexistence of multiple traffic velocities at the same density.

For the more complex KKW-1 dynamics, PI-NCA also retains the broad distribution of traffic states and the transition from high-velocity states at lower local densities to lower-velocity congested states as density increases. PI-NCA also recovers the overall shape of the binned local flow-density relation. \ref{appendix:learned_stoch_behavior} provides a complementary analysis of the categorical velocity distributions predicted during the simulation, in which we examine the learned velocity probabilities as a function of local density. 

At the global scale, we run PI-NCA simulations on a road of length $L=1000$ using the same condensed and homogeneous initial conditions as the corresponding reference simulations. For each density and initialization, we average the global traffic quantities over 1000 time steps and across 5 independent realizations.

For NaSch dynamics, Figure~\ref{fig:global_fd_pinca_nasch} shows that the stochastic PI-NCA closely reproduces the reference model's global density--flow relation for both initial traffic configurations. In particular, it captures the increase in flow toward the maximum-flow region, the separation between the condensed and homogeneous initializations, and their subsequent convergence as the density increases. The relative $L^2$ errors between PI-NCA and reference flow curves, defined in~\ref{appendix:metrics}, are $1.61\%$ for the condensed initialization and $1.45\%$ for the homogeneous initialization, indicating a close similarity across the considered density range.

For the KKW-1 dynamics, similarity is also observed at the global scale. Figure~\ref{fig:global_fd_pinca_kkw1} shows that PI-NCA reproduces the dependence of the KKW-1 density--flow relation on the initial traffic configurations. In particular, the distinct condensed and homogeneous states observed for KKW-1 reappear around the maximum-flow region with the spiky behavior, followed by convergence toward a common congested state at higher densities. 
The corresponding relative $L^2$ errors are $4.70\%$ and $5.16\%$ for the condensed and homogeneous cases, respectively. 

Overall, these results show that the local stochastic dynamics learned by PI-NCA recover characteristic traffic behavior at both local and global scales during autoregressive rollout.

%%%%%%%%%%%%%%%%%%%%%%%%%%%%%%%%%%%%%%%%%%%%%%%%%%%%%%%%%%%%%%%%%%
\section{Summary and Conclusions}
\label{sec:conclusion}

This work investigates Neural Cellular Automata and their use for learning deterministic and stochastic traffic dynamics from observed state trajectories. We consider two formulations: an NCA that directly predicts the traffic state and a Physics-Informed NCA (PI-NCA) that learns vehicle velocities while explicitly updating their positions. We evaluate the approaches using two traffic CA models: NaSCh and the more complex KKW-1 models.

Overall, the results show that PI-NCA provides more accurate and physically consistent predictions than the direct state prediction model. The largest improvements appear with the KKW-1 model, as the NCA for these dynamics shows the highest errors, which decrease when physics is considered. The learned local rule generalizes across different initializations, road lengths, and vehicle numbers, even for configurations not seen during training. For stochastic dynamics, PI-NCA also reproduces the characteristic local density--flow behavior of the reference models, and the learned velocity distributions show a shift from high-speed free flow to congested traffic as local density increases.

These findings suggest that combining learned local interaction rules with known physical structure is a promising approach for data-driven traffic modeling. Future work should extend the framework to more realistic settings, including higher and continuous velocities, open boundaries, multi-lane and heterogeneous traffic.

	%%%%%%%%%%%%%%%%%%%%%%%%%%%%%%%%%%%%%%%%%%%%%%%%%%%%%%%%%%%%%%%%%%
	\section*{ACKNOWLEDGMENTS}
	
	The authors would like to thank the Wallenberg AI, Autonomous Systems and Software Program (WASP) funded by the Knut and Alice Wallenberg Foundation for support.

\bibliographystyle{cas-model2-names}

\bibliography{references}

    \appendix

\newpage

\section{Evaluation Metrics}
\label{appendix:metrics}

We compute all reported metrics over complete rollout trajectories. Let $\mathbf{s}_{\tau}\in\{-1,0,\ldots,v_{\max}\}^{L}$ denote the reference traffic state at time step $\tau$, and let $\hat{\mathbf{s}}_{\tau}$ denote the corresponding model prediction. Furthermore, let $\mathcal{C}_{\tau} = \{\,i : s_{\tau}(i)\ge0\,\}$ be the set of occupied cells at time $\tau$, and let $n_{\tau}=|\mathcal{C}_{\tau}|$ denote the number of vehicles.

The following metrics are evaluated for every rollout and subsequently averaged
over all test trajectories and independent training runs.

\subsection*{Mean Absolute Error}
The Mean Absolute Error (MAE) measures the average absolute difference between predicted and reference cell values,

\begin{equation}
\mathrm{MAE} = \frac{1}{(T+1)L}\sum_{\tau=0}^{T}\sum_{i\in \Omega}\left|s_{\tau}(i)-\hat{s}_{\tau}(i)\right|.
\end{equation}

\subsection*{Mean Squared Error}
The Mean Squared Error (MSE) is defined as

\begin{equation}
\mathrm{MSE} = \frac{1}{(T+1)L}\sum_{\tau=0}^{T}\sum_{i\in \Omega}\left(s_{\tau}(i)-\hat{s}_{\tau}(i)\right)^2.
\end{equation}

\subsection*{Car Mean Absolute Error}
To evaluate prediction accuracy only on vehicles, the Car MAE is computed over occupied cells,

\begin{equation}
\mathrm{CarMAE} = \frac{1}{T+1}\sum_{\tau=0}^{T}\frac{1}{n_{\tau}}\sum_{i\in\mathcal{C}_{\tau}}\left|s_{\tau}(i)-\hat{s}_{\tau}(i)\right|.
\end{equation}

\subsection*{Occupancy Accuracy}
Occupancy Accuracy (OA) measures the fraction of cells whose occupancy state (empty or occupied) is predicted correctly,

\begin{equation}
\mathrm{OA} = \frac{1}{(T+1)L}\sum_{\tau=0}^{T}\sum_{i\in \Omega}\mathbf{1}\left[(s_{\tau}(i)\ge0) = (\hat{s}_{\tau}(i)\ge0)\right].
\end{equation}

\subsection*{Conservation Error}
The Conservation Error (CE) quantifies deviations in the total number of vehicles,

\begin{equation}
\mathrm{CE} = \frac{1}{T+1}\sum_{\tau=0}^{T}\left|n_{\tau}-\hat n_{\tau}\right|,
\end{equation}

where $\hat n_{\tau} = \left|\left\{i:\hat{s}_{\tau}(i)\ge0 \right\}\right|$.

\subsection*{Relative Conservation Error}
The Relative Conservation Error (RCE) normalizes the conservation error by the reference vehicle count,

\begin{equation}
\mathrm{RCE} = \frac{1}{T+1}\sum_{\tau=0}^{T}\frac{|n_{\tau}-\hat n_{\tau}|}{\max(n_{\tau},1)}.
\end{equation}

\subsection*{Total Variation Distance}
The Total Variation (TV) distance measures the agreement between the predicted and the reference next-velocity probability distributions. Let $P_{\mathrm{ref}}$ denote the reference distribution, and $P_{\mathrm{PI-NCA}}$ the corresponding predicted distribution. The TV distance is defined as
\begin{equation}
    D_{TV} = \frac{1}{2}\sum_{k=0}^{v_{\text{max}}}\lvert P_{\mathrm{ref}} \left(v_t(i)=k \mid \mathbf{s}_{t-1}\right) - P_{\mathrm{PI-NCA}} \left(v_t(i)=k \mid \mathbf{s}_{t-1}\right) \rvert.
\end{equation}
The metric ranges from 0 to 1, where 0 indicates identical probability distributions.

\subsection{Relative $L^2$ Error}
The Relative $L^2$ Error measures the agreement between the global fundamental diagrams of the learned dynamics using stochastic PI-NCA and the reference traffic flow models. It is defined as
\begin{equation}
    \epsilon_{L^2} = \frac{\left|\mathbf{q}_{\mathrm{PI-NCA}}-\mathbf{q}_{\mathrm{ref}}\right|_2}{\left|\mathbf{q}_{\mathrm{ref}}\right|_2},
\end{equation}
where $\mathbf{q}_{\mathrm{ref}}$ and $\mathbf{q}_{\mathrm{ref}}$ denote the global flow curves of PI-NCA and corresponding reference model, respectively, evaluated at the same global densities.

\section{KKW-1 Reference Dynamics}
\label{appendix:kkw1_reference_dynamics}
To evaluate if the proposed NCA formulations can learn more complex traffic dynamics other than generated by the NaSch model, we also consider the KKW-1 CA proposed by Kerner, Klenov and Wolf \cite{kerner2002cellular}. We use an adapted version of KKW-1, with road discretization as introduced in Section~\ref{subsec:extension_kkw1}.

For a vehicle occupying cell $i$ at time $t-1$, let $v_{t-1}(i)$ be its velocity, $v_{l,t-1}(i)$ the velocity of the leading vehicle, and $g_{t-1}(i)$ the available gap. The corresponding safe velocity is
\begin{equation}
    v_{s,t}(i) = \frac{g_{t-1}(i)}{\tau},
\end{equation}
where $\tau$ is the time-step duration. The synchronization distance is
\begin{equation}
    D_{t-1}(i) = d + kv_{t-1}(i)\tau,
\end{equation}
where $d$ is the vehicle length, and $k$ is the synchronization parameter. This parameter determines the range over which the vehicle adapts to the leading vehicle.
The control velocity is given by
\begin{equation}
    v_{c,t}(i) =
    \begin{cases}
        v_{t-1}(i) + a\tau,  &  g_{t-1}(i) > D_{t-1}(i) - d, \\
        v_{t-1}(i) + a\tau ~\mathrm{sgn}(v_{l, t-1}(i) - v_{t-1}(i)),  &  g_{t-1}(i) \leq D_{t-1}(i) - d,
    \end{cases}
\end{equation}
where $a$ is the acceleration, and $\mathrm{sgn(\cdot)}$ is the sign function.
The deterministic velocity is obtained in the following way
\begin{equation}
    v_t^{\mathrm{det}}(i) = \mathrm{max}\left(0, \mathrm{min}\left(v_{\mathrm{max}}, v_{s,t}(i), v_{c,t}(i)\right)\right)
\end{equation}
where $v_{\max}$ is the maximum velocity. For the stochastic dynamics,
\begin{equation}
    v_t(i) = \mathrm{max}\left(0, \mathrm{min}\left(v_t^{\mathrm{det}}(i) + a\tau \eta_t(i), v_{t-1}(i) + a\tau, v_{\mathrm{max}}, v_{s,t}(i)\right)\right)
\end{equation}
where $\eta_t(i)$ represents the stochastic acceleration or deceleration and is given by
\begin{equation}
    \eta_t(i) = 
    \begin{cases}
        -1,  &  r < p_b, \\
        ~~1,  &  p_b \leq r < p_b + p_a, \\
        ~~0,  &  \mathrm{otherwise},
    \end{cases}
\end{equation}
where $r\sim U(0,1)$ is uniformly sampled from $[0, 1]$, and $p_b$ and $p_a$ are the probabilities of stochastic deceleration and acceleration, respectively. These probabilities are
\begin{equation}
    p_b = 
    \begin{cases}
        p_0,  &  v_{t-1}(i) = 0, \\
        p,  &  v_{t-1}(i) > 0,
    \end{cases}
    \qquad
    p_a =
    \begin{cases}
        p_{a1},  &  v_{t-1}(i) < v_p, \\
        p_{a2},  &  v_{t-1}(i) \geq v_p,
    \end{cases}
\end{equation}
where $p_0$ and $p$ are the deceleration probabilities for stopped and moving vehicles, respectively; $p_{a1}$ and $p_{a2}$ are the acceleration probabilities below and above the velocity threshold $v_p$.

Finally, the vehicle moves accordingly, as in the NaSch model, to
\begin{equation}
    j = (i + v_t(i)) \bmod L,
\end{equation}
where $j$ is the destination cell and $L$ is the road length. The complete traffic state at time $t$ is then reconstructed as
\begin{equation}
    s_t(j) = 
    \begin{cases}
        v_t(i), & \text{if cell } j \text{ is occupied by the vehicle moving from cell } i, \\
        -1, & \text{otherwise}.
    \end{cases}
\end{equation}
where $s_t(j)$ represents the state of cell $j$ at time $t$, and -1 denotes an empty cell.

Table \ref{tab:kkw1_parameters} lists the adapted KKW-1 parameters for the experiments.

\begin{table}[h!]
    \centering
    \small
    \setlength{\tabcolsep}{15pt}
    \renewcommand{\arraystretch}{1.25}
    \caption{Parameters used for the adapted KKW-1 model.}
    \label{tab:kkw1_parameters}
    
    \begin{tabular}{@{}llc@{}}
    \toprule

    \textbf{Parameter}  &  \textbf{Description}  &  \textbf{Value} \\
    \hline
    $d$  &  Vehicle length  &  1 \\
    $v_{\mathrm{max}}$  &  Maximum vehicle velocity  &  5  \\
    $a$  &  Vehicle acceleration  &  1  \\
    $\tau$  &  Time-step duration  &  1  \\
    $k$  &  Synchronization parameter  &  2.04  \\
    $v_p$  &  Velocity threshold for random acceleration  &  2.33  \\
    $p_0$  &  Deceleration probability for stopped vehicles  &  0.425  \\
    $p$  &  Deceleration probability for moving vehicles  &  0.04  \\
    $p_{a1}$  &  Acceleration probability below $v_p$  &  0.20  \\
    $p_{a2}$  &  Acceleration probability at or above $v_p$  &  0.052 \\
    \bottomrule
    \end{tabular}
\end{table}

\newpage

\section{Additional Performance Tables}
\label{appendix:additional_performance_tables}

We provide the detailed deterministic prediction results for the NCA and PI-NCA formulations evaluated against both the NaSch and KKW-1 reference dynamics. All values are reported as the mean and standard deviation over 5 independent training runs. Table \ref{tab:appendix_global} complements the overall results reported in Table \ref{tab:global} by providing the additional MSE, occupancy accuracy (OA), and relative conservation error (RCE) metrics.
Tables~\ref{tab:initcase},~\ref{tab:L}, and~\ref{tab:density} provide the numerical MAE and conservation error (CE) results underlying Figures~\ref{fig:results_initcase},~\ref{fig:results_L}, and~\ref{fig:results_num_cars}, respectively, reporting performance across the different initialization strategies, road lengths, and vehicle counts considered in the test set.

\begin{table}[h!]
\centering
\caption{Overall prediction performance over the complete NaSch and KKW-1 test sets.}
\label{tab:appendix_global}

\small
\setlength{\tabcolsep}{4pt}
\renewcommand{\arraystretch}{1.25}

\begin{tabular}{@{}llccc@{}}
\toprule
\textbf{Traffic CA}
& \textbf{Model}
& \textbf{MSE}
& \textbf{OA}
& \textbf{RCE} \\
\midrule

\multirow{2}{*}{NaSch}
& NCA
& $0.0105 \pm 0.0089$
& $0.9995 \pm 0.0003$
& $0.0004 \pm 0.0002$ \\

& PI-NCA
& $0.0002 \pm 0.0003$
& $0.9999 \pm 0.0000$
& $0.0000 \pm 0.0000 $ \\

\midrule

\multirow{2}{*}{KKW-1}
& NCA
& $0.7810 \pm 0.1274$
& $0.9472 \pm 0.0071$
& $0.0194 \pm 0.0050$ \\

& PI-NCA
& $0.0655 \pm 0.0320$
& $0.9946 \pm 0.0028$
& $0.00002 \pm 0.00005 $ \\

\bottomrule
\end{tabular}
\end{table}

\begin{table}[h!]
\centering
\caption{Prediction performance across the different initialization strategies.}
\label{tab:initcase}

\small
\setlength{\tabcolsep}{4pt}
\renewcommand{\arraystretch}{1.25}

\begin{tabular}{@{}lllcc@{}}
\toprule
\textbf{Init.}
& \textbf{Traffic CA}
& \textbf{Model}
& \textbf{MAE}
& \textbf{CE} \\
\midrule

\multirow{4}{*}{$I_1$}  
& \multirow{2}{*}{NaSch}
& NCA &  $0.0000 \pm 0.0000$ & $0.0000 \pm 0.0000 $ \\
& & PI-NCA & $0.0000 \pm 0.0000$ & $0.0000 \pm 0.0000 $ \\
\cmidrule(lr){2-5}
& \multirow{2}{*}{KKW-1}
& NCA &  $0.0386 \pm 0.0509$ & $0.1350 \pm 0.1154$ \\
& & PI-NCA & $0.0000 \pm 0.0000$ & $0.0000 \pm 0.0000$ \\

\midrule
\multirow{4}{*}{$I_2$}
& \multirow{2}{*}{NaSch}
& NCA &  $0.0000 \pm 0.0000$ & $0.0000 \pm 0.0000 $ \\
& & PI-NCA & $0.0000 \pm 0.0000$ & $0.0000 \pm 0.0000 $ \\
\cmidrule(lr){2-5}
& \multirow{2}{*}{KKW-1}
& NCA &  $0.1193 \pm 0.0269$ & $0.2766 \pm 0.1571$ \\
& & PI-NCA & $0.0000 \pm 0.0000$ & $0.0000 \pm 0.0000 $ \\

\midrule
\multirow{4}{*}{$I_3$}
& \multirow{2}{*}{NaSch}
& NCA &  $0.0058 \pm 0.0079$ & $0.0127 \pm 0.0284$ \\
& & PI-NCA & $0.0000 \pm 0.0000$ & $0.0000 \pm 0.0000 $ \\
\cmidrule(lr){2-5}
& \multirow{2}{*}{KKW-1}
& NCA &  $0.2037 \pm 0.0568$ & $0.3369 \pm 0.1671$ \\
& & PI-NCA & $0.0036 \pm 0.0081$ & $0.0000 \pm 0.0000 $ \\

\midrule
\multirow{4}{*}{$I_4$}  
& \multirow{2}{*}{NaSch}
& NCA &  $0.0000 \pm 0.0000$ & $0.0000 \pm 0.0000 $ \\
& & PI-NCA & $0.0000 \pm 0.0000$ & $0.0000 \pm 0.0000 $ \\
\cmidrule(lr){2-5}
& \multirow{2}{*}{KKW-1}
& NCA &  $0.0281 \pm 0.0261$ & $0.0268 \pm 0.0384$ \\
& & PI-NCA & $0.0000 \pm 0.0000$ & $0.0000 \pm 0.0000$ \\

\midrule
\multirow{4}{*}{$I_5$}  
& \multirow{2}{*}{NaSch}
& NCA &  $0.0004 \pm 0.0004$ & $0.0042 \pm 0.0057$ \\
& & PI-NCA & $0.0000 \pm 0.0000$ & $0.0000 \pm 0.0000 $ \\
\cmidrule(lr){2-5}
& \multirow{2}{*}{KKW-1}
& NCA &  $0.2011 \pm 0.0525$ & $0.2978 \pm 0.0834$ \\
& & PI-NCA & $0.0117 \pm 0.0142$ & $0.0000 \pm 0.0000$ \\

\midrule
\multirow{4}{*}{$I_6$}
& \multirow{2}{*}{NaSch}
& NCA &  $0.0071 \pm 0.0059$ & $0.0258 \pm 0.0166$ \\
& & PI-NCA & $0.0004 \pm 0.0006$ & $0.0011 \pm 0.0024$ \\
\cmidrule(lr){2-5}
& \multirow{2}{*}{KKW-1}
& NCA &  $0.5784 \pm 0.0760$ & $0.7780 \pm 0.1103$ \\
& & PI-NCA & $0.0921 \pm 0.0410$ & $0.0010 \pm 0.0023$ \\  

\bottomrule
\end{tabular}
\end{table}

\begin{table}[h!]
\centering
\caption{Prediction performance for different road lengths $L$. The values $L\in\{50,150\}$ are out-of-distribution.}
\label{tab:L}

\small
\setlength{\tabcolsep}{4pt}
\renewcommand{\arraystretch}{1.25}

\begin{tabular}{@{}lllcc@{}}
\toprule
\textbf{$L$}
& \textbf{Traffic CA}
& \textbf{Model}
& \textbf{MAE}
& \textbf{CE} \\
\midrule

\multirow{4}{*}{50}
& \multirow{2}{*}{NaSch}
& NCA &  $0.0025 \pm 0.0008$ & $0.0160 \pm 0.0076$ \\
& & PI-NCA & $0.0002 \pm 0.0003$ & $0.0005 \pm 0.0012$ \\
\cmidrule(lr){2-5}
& \multirow{2}{*}{KKW-1}
& NCA &  $0.4072 \pm 0.0675$ & $0.7136 \pm 0.2020$ \\
& & PI-NCA & $0.0421 \pm 0.0235$ & $0.0005 \pm 0.0012 $ \\

\midrule
\multirow{4}{*}{100}
& \multirow{2}{*}{NaSch}
& NCA &  $0.0031 \pm 0.0032$ & $0.0032 \pm 0.0047$ \\
& & PI-NCA & $0.0000 \pm 0.0000$ & $0.0000 \pm 0.0000 $ \\
\cmidrule(lr){2-5}
& \multirow{2}{*}{KKW-1}
& NCA &  $0.1279 \pm 0.0062$ & $0.1321 \pm 0.0322$ \\
& & PI-NCA & $0.0069 \pm 0.0050$ & $0.0000 \pm 0.0000 $ \\
 
\midrule
\multirow{4}{*}{150}  
& \multirow{2}{*}{NaSch}
& NCA &  $0.0009 \pm 0.0018$ & $0.0021 \pm 0.0047$ \\
& & PI-NCA & $0.0000 \pm 0.0000$ & $0.0000 \pm 0.0000 $ \\
\cmidrule(lr){2-5}
& \multirow{2}{*}{KKW-1}
& NCA &  $0.0495 \pm 0.0152$ & $0.0797 \pm 0.0369$ \\
& & PI-NCA & $0.0047 \pm 0.0044$ & $0.0000 \pm 0.0000$ \\ 

\bottomrule
\end{tabular}
\end{table}

\begin{table}[h!]
\centering
\caption{Prediction performance for seen and unseen numbers of vehicles. The values $\{8,12,17\}$ are out-of-distribution.}
\label{tab:density}

\small
\setlength{\tabcolsep}{4pt}
\renewcommand{\arraystretch}{1.25}

\begin{tabular}{@{}lllcc@{}}
\toprule
\textbf{$N$}
& \textbf{Traffic CA}
& \textbf{Model}
& \textbf{MAE}
& \textbf{CE} \\
\midrule

\multirow{4}{*}{8}
& \multirow{2}{*}{NaSch}
& NCA &  $0.0002 \pm 0.0005$ & $0.0000 \pm 0.0000 $ \\
& & PI-NCA & $0.0000 \pm 0.0000$ & $0.0000 \pm 0.0000 $ \\
\cmidrule(lr){2-5}
& \multirow{2}{*}{KKW-1}
& NCA &  $0.0522 \pm 0.0140$ & $0.0479 \pm 0.0131$ \\
& & PI-NCA & $0.0051 \pm 0.0050$ & $0.0010 \pm 0.0023 $ \\

\midrule
\multirow{4}{*}{10}
& \multirow{2}{*}{NaSch}
& NCA &  $0.0008 \pm 0.0015$ & $0.0011 \pm 0.0024$ \\
& & PI-NCA & $0.0000 \pm 0.0000$ & $0.0000 \pm 0.0000 $ \\
\cmidrule(lr){2-5}
& \multirow{2}{*}{KKW-1}
& NCA &  $0.0870 \pm 0.0332$ & $0.0904 \pm 0.0433$ \\
& & PI-NCA & $0.0090 \pm 0.00711$ & $0.0000 \pm 0.0000 $ \\
 
\midrule
\multirow{4}{*}{12}  
& \multirow{2}{*}{NaSch}
& NCA &  $0.0016 \pm 0.0015$ & $0.0053 \pm 0.0092 $ \\
& & PI-NCA & $0.0000 \pm 0.0000$ & $0.0000 \pm 0.0000 $ \\
\cmidrule(lr){2-5}
& \multirow{2}{*}{KKW-1}
& NCA &  $0.1427 \pm 0.0505$ & $0.1288 \pm 0.0197$ \\
& & PI-NCA & $0.0119 \pm 0.0051$ & $0.0000 \pm 0.0000$ \\
 
\midrule
\multirow{4}{*}{15}  
& \multirow{2}{*}{NaSch}
& NCA &  $0.0031 \pm 0.0056$ & $0.0063 \pm 0.0058 $ \\
& & PI-NCA & $0.00002 \pm 0.00002$ & $0.0000 \pm 0.0000 $ \\
\cmidrule(lr){2-5}
& \multirow{2}{*}{KKW-1}
& NCA &  $0.2653 \pm 0.0263$ & $0.3775 \pm 0.0922$ \\
& & PI-NCA & $0.0191 \pm 0.0161$ & $0.0000 \pm 0.0000$ \\

\midrule
\multirow{4}{*}{17}  
& \multirow{2}{*}{NaSch}
& NCA &  $0.0042 \pm 0.0037$ & $0.0149 \pm 0.0094$ \\
& & PI-NCA & $0.0002 \pm 0.0004$ & $0.0000 \pm 0.0000 $ \\
\cmidrule(lr){2-5}
& \multirow{2}{*}{KKW-1}
& NCA &  $0.3199 \pm 0.0548$ & $0.5904 \pm 0.1988$ \\
& & PI-NCA & $0.0281 \pm 0.0166$ & $0.0000 \pm 0.0000$ \\

\midrule
\multirow{4}{*}{20}
& \multirow{2}{*}{NaSch}
& NCA &  $0.0034 \pm 0.0025$ & $0.0150 \pm 0.0069$ \\
& & PI-NCA & $0.0001 \pm 0.0002$ & $0.0011 \pm 0.0024$ \\
\cmidrule(lr){2-5}
& \multirow{2}{*}{KKW-1}
& NCA &  $0.3021 \pm 0.0441$ & $0.6159 \pm 0.1915$ \\
& & PI-NCA & $0.0343 \pm 0.0180$ & $0.0000 \pm 0.0000$ \\  

\bottomrule
\end{tabular}
\end{table}

\newpage

\section{Learned Stochastic Behavior}
\label{appendix:learned_stoch_behavior}

\begin{figure}[h!]
    \centering
    \includegraphics[width=1.0\linewidth]{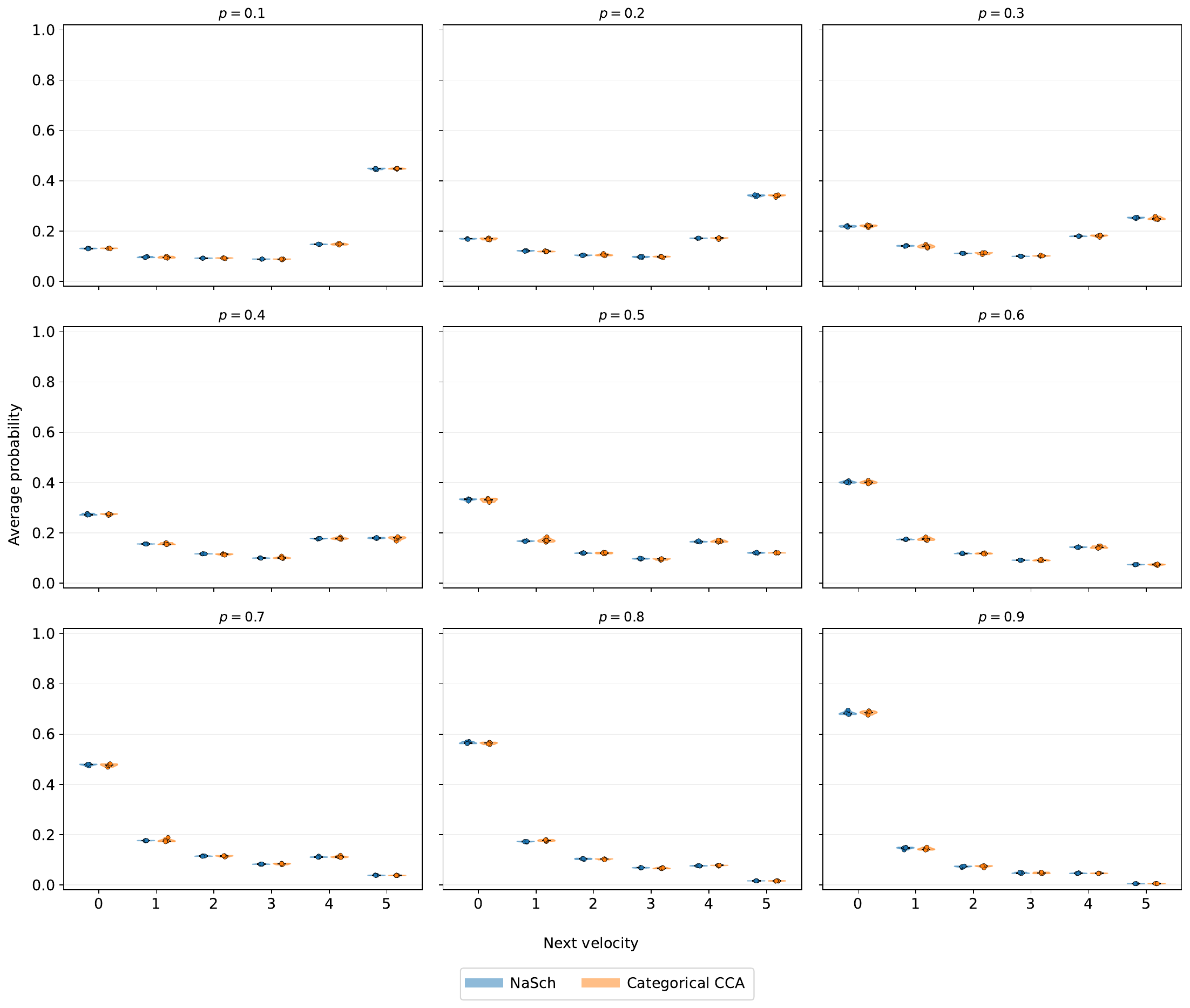}
    \caption{Average probability assigned to each next velocity class for $p\in\{0.1, 0.2, \dots\}$, comparing the reference NaSch distribution with the learned PI-NCA distribution.}
    \label{fig:all_p_slow_pinca}
\end{figure}

\begin{figure}[h!]
    \centering
    \includegraphics[width=0.7\linewidth]{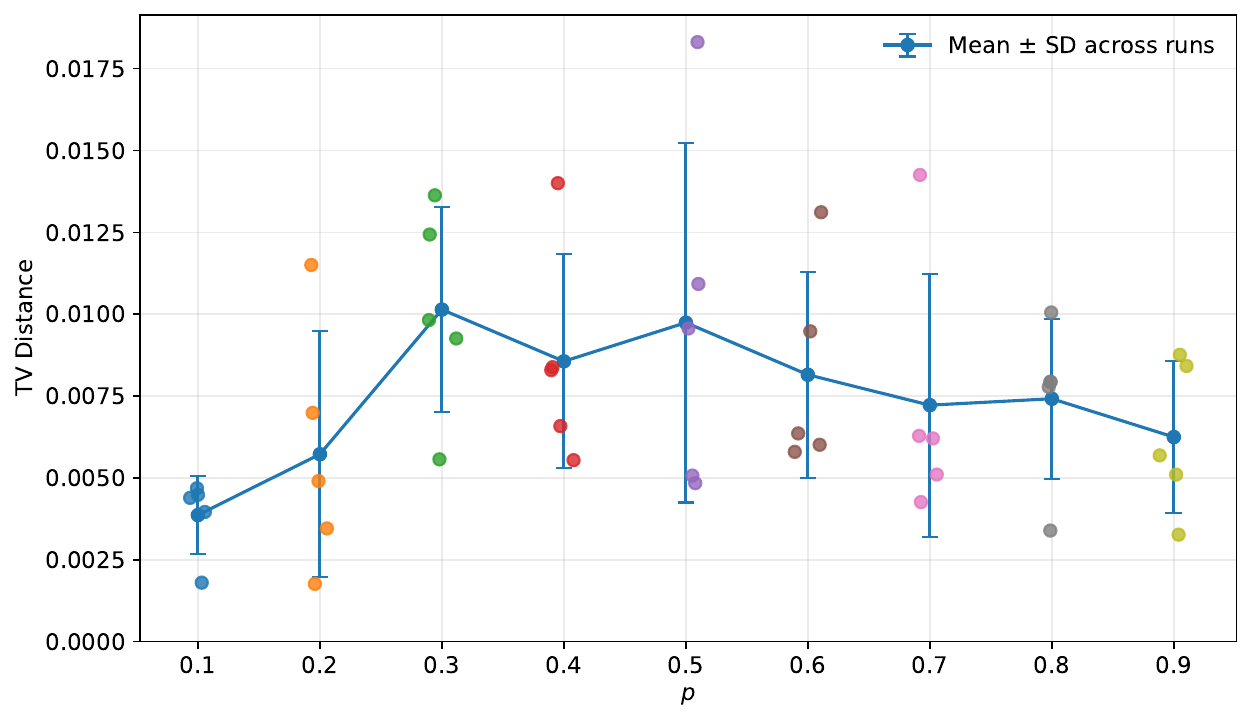}
    \caption {Total variation distance between the reference and predicted distributions. Points denote independent training runs.}
    \label{fig:tvd_all_pslow}
\end{figure}

We conduct additional experiments to show that the learned model recovers the NaSch stochastic transition rule. For the NaSch model, the stochastic mechanism is known explicitly, allowing a direct comparison with the learned categorical distributions. Figure~\ref{fig:all_p_slow_pinca} reports the average probability assigned to each next velocity class for randomization probabilities $p\in\{0.1, 0.2, \dots, 0.9\}$. For all the ranges of $p$, the stochastic PI-NCA distributions closely follow the corresponding NaSch distributions. Additionally, Figure~\ref{fig:tvd_all_pslow} quantifies the Total Variation (TV) distance between the two distributions for each $p$. The mean TV distance remains small throughout the tested range of randomization probabilities.

\begin{figure}[h!]
     \centering
     \begin{subfigure}[b]{0.48\textwidth}
         \centering
         \includegraphics[width=\textwidth]{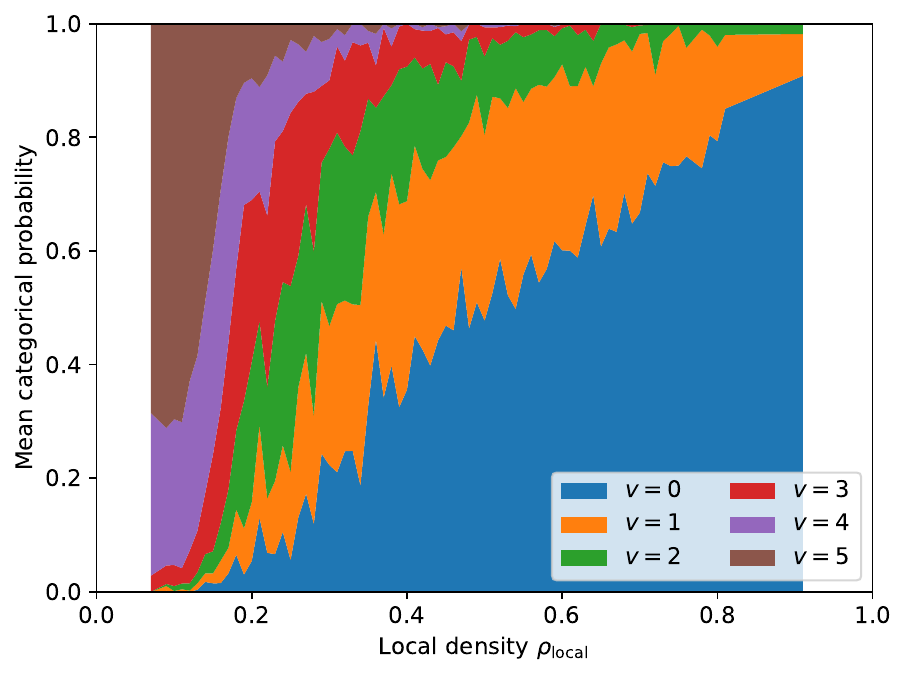}
         \caption{NaSch probabilities}
         \label{fig:stacked_prob_pinca_nasch}
     \end{subfigure}
     %\hfill
     \begin{subfigure}[b]{0.48\textwidth}
         \centering
         \includegraphics[width=\textwidth]{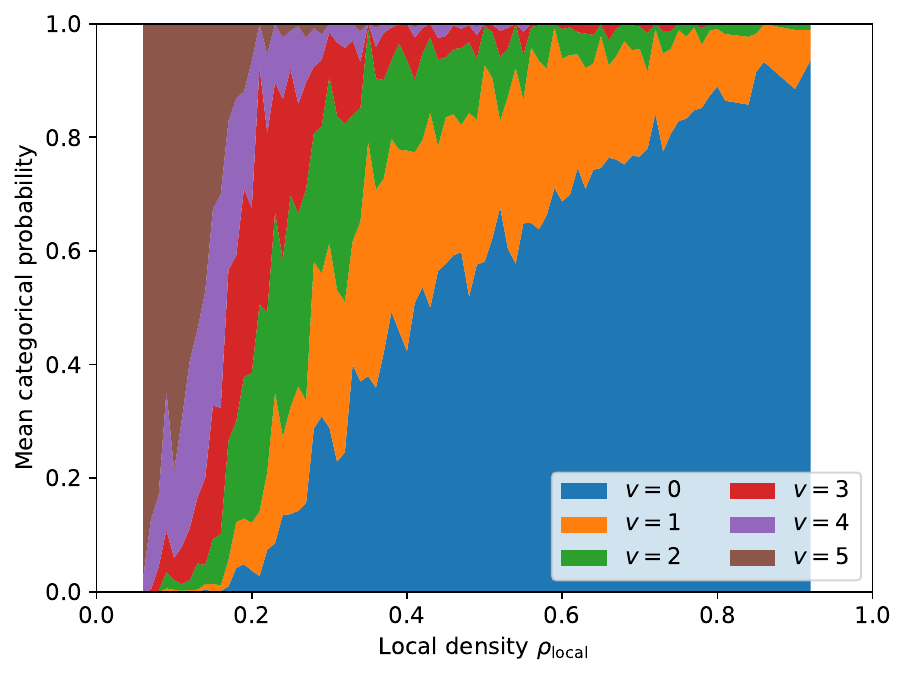}
         \caption{KKW-1 probabilities}
         \label{fig:stacked_prob_pinca_kkw1}
     \end{subfigure}
\caption{Mean categorical next velocity probabilities predicted by PI-NCA as a function of local density for (a) NaSch and (b) KKW-1, respectively.}
\label{fig:stacked_probs_pinca}
\end{figure}

To further examine the stochastic behavior learned by PI-NCA, Figures~\ref{fig:stacked_prob_pinca_nasch} and \ref{fig:stacked_prob_pinca_kkw1} provide a complementary view for the NaSch and KKW-1 dynamics, respectively. The results are obtained from the same single PI-NCA simulation used to construct the corresponding local fundamental diagrams presented in Figures~\ref{fig:local_fd_pinca_nasch} and~\ref{fig:local_fd_pinca_kkw1}. For each occupied cell, the model predicts a categorical distribution over the possible next velocities. We aggregate these probability distributions using the same detector locations and temporal windows as the local fundamental diagrams, and group them by local density. As density increases, probability mass moves from higher to lower velocity classes for both NaSch and KKW-1 dynamics. In both the local fundamental diagrams and the stacked probabilities, we observe that as local density increases, the learned velocity distribution shifts from faster to slower vehicle movement, reducing local flow. 

% 

% \section{Additional Fundamental Diagrams}
% \label{appendix:additional_fundamental_diagrams}

%\begin{thebibliography}{00}

%% For numbered reference style

%   \textit{\LaTeX: a document preparation system},
%   Addison Wesley, Massachusetts,

% \end{thebibliography}
\end{document}